\documentclass[sigconf]{acmart}
\makeatletter
\global\@ACM@balancefalse
\makeatother

\usepackage{multirow}
\usepackage{colortbl}
\usepackage{placeins}
\usepackage{balance}

\usepackage{bm}

\def\eqref#1{equation~\ref{#1}}

\def\1{\bm{1}}

\DeclareMathAlphabet{\mathsfit}{\encodingdefault}{\sfdefault}{m}{sl}
\SetMathAlphabet{\mathsfit}{bold}{\encodingdefault}{\sfdefault}{bx}{n}

\newcommand{\ie}{\textit{i.e.,~}}

\usepackage[inline]{enumitem}
\setlist[description]{leftmargin=\parindent,labelindent=\parindent, font=\normalfont\itshape}

\usepackage{multirow}
\usepackage{tabularx}
\usepackage{cleveref} 

\crefname{section}{\S}{\S} 
\crefname{subsection}{\S}{\S} 
\newif\ifshowpendingexperiments
\showpendingexperimentsfalse
\newcommand{\pendingexperiment}[2]{%
  \ifshowpendingexperiments
    \par\medskip
    \noindent\fbox{\parbox{\dimexpr\linewidth-2\fboxsep-2\fboxrule\relax}{%
      \textbf{Draft experiment placeholder---#1.} #2}}%
    \par\medskip
  \fi
}

\setcopyright{none}
\renewcommand\footnotetextcopyrightpermission[1]{}
\acmConference[KDD '27]{ACM SIGKDD Conference on Knowledge Discovery and Data Mining}{2027}{San Jose, CA, USA}

\begin{document}

\title{Understanding Semantic IDs: From Item Representation to Item Selection in Generative Recommendation}

\author{Junting Wang}
\email{junting3@illinois.edu}
\affiliation{%
  \institution{University of Illinois Urbana-Champaign}
  \city{Urbana}
  \state{Illinois}
  \country{USA}
}

\author{Xinrui He}
\email{xhe33@illinois.edu}
\affiliation{%
  \institution{University of Illinois Urbana-Champaign}
  \city{Urbana}
  \state{Illinois}
  \country{USA}
}

\author{Yunzhe Li}
\email{yunzhel2@illinois.edu}
\affiliation{%
  \institution{University of Illinois Urbana-Champaign}
  \city{Urbana}
  \state{Illinois}
  \country{USA}
}

\author{Hari Sundaram}
\email{hs1@illinois.edu}
\affiliation{%
  \institution{University of Illinois Urbana-Champaign}
  \city{Urbana}
  \state{Illinois}
  \country{USA}
}

\begin{abstract}
Semantic IDs (SIDs) are now a central component of generative recommendation.  Current SID-based systems assign three roles to the same token sequence. Shared prefixes are intended to organize related items, the complete SID identifies an individual item, and each generated token narrows the items that can still be returned. We systematically investigate SIDs from item encoding and SID construction to autoregressive generation and final recommendation. We examine how SID construction changes item representations and how those changes affect generation. Across three Amazon domains and eight SID constructions, SID neighborhoods recover only 32.2\% of the encoder's ten nearest neighbors on average. Alternative item descriptions still retrieve the corresponding item first in 99.57\% of controlled cases, yet change 38.4\% of exact SIDs. These results show that SIDs retain broad organization but lose much of the encoder's fine local structure, while their exact tokens are not determined by item meaning alone. This loss becomes consequential during generation. After the final semantic token, TIGER retains only 29.9\% of held-out targets that were plausible recommendations before SID filtering. Motivated by these findings, we propose Item-Supported Decoding (ISD), a lightweight inference-time method that allows a user-specific item ranking to support corresponding SID prefixes before beam search discards them. The same ranking then orders the generated items. ISD requires no additional parameters or retraining of the SID constructor or decoder. We empirically show that ISD improves NDCG@10 over the corresponding SID backbone in every evaluated setting, with relative gains of up to 31.2\%. Our results show that SIDs provide useful coarse item organization, but their fine boundaries should not alone determine which items remain available during generation. Code and experimental artifacts are available at \url{https://anonymous.4open.science/r/ISD-CF7C}.

\end{abstract}


\maketitle

\section{Introduction}
\label{sec:introduction}

Generative recommenders increasingly use Semantic IDs (SIDs) to represent items~\cite{rajput2023recommender,wang2024letter,liu2024etegrec,hou2025rpg,fu2026diger}. They use codebooks to convert item content into short sequences of discrete tokens. The motivation is compelling: content guides how items are tokenized, while the shared codebook allows related products to share tokens or prefixes. Therefore, during training, the interaction data can reinforce common parts of the prediction instead of teaching the model to recognize every item as an unrelated label.

Existing SID recommenders use shared tokens to encode common semantic or collaborative structure~\cite{rajput2023recommender,wang2024letter}, while treating the complete token sequence as the item identifier to be generated~\cite{rajput2023recommender,liu2024etegrec,fu2026diger}. This places two demands on the same representation. The shared tokens should organize related items so that they can benefit from common structure, while the complete sequence must uniquely identify the item. During decoding, autoregressive SID recommenders generate the prediction sequence using constrained beam search~\cite{rajput2023recommender,wang2024letter,liu2024etegrec}. Autoregressive generation therefore gives the SID a third role in item selection. At each decoding step, the predicted token retains only the items with that prefix and removes the rest. Therefore, the distinctions introduced to identify individual items also become intermediate decisions about which items remain eligible for recommendation.

\begin{figure}[t]
  \centering
  \includegraphics[width=0.48\textwidth]{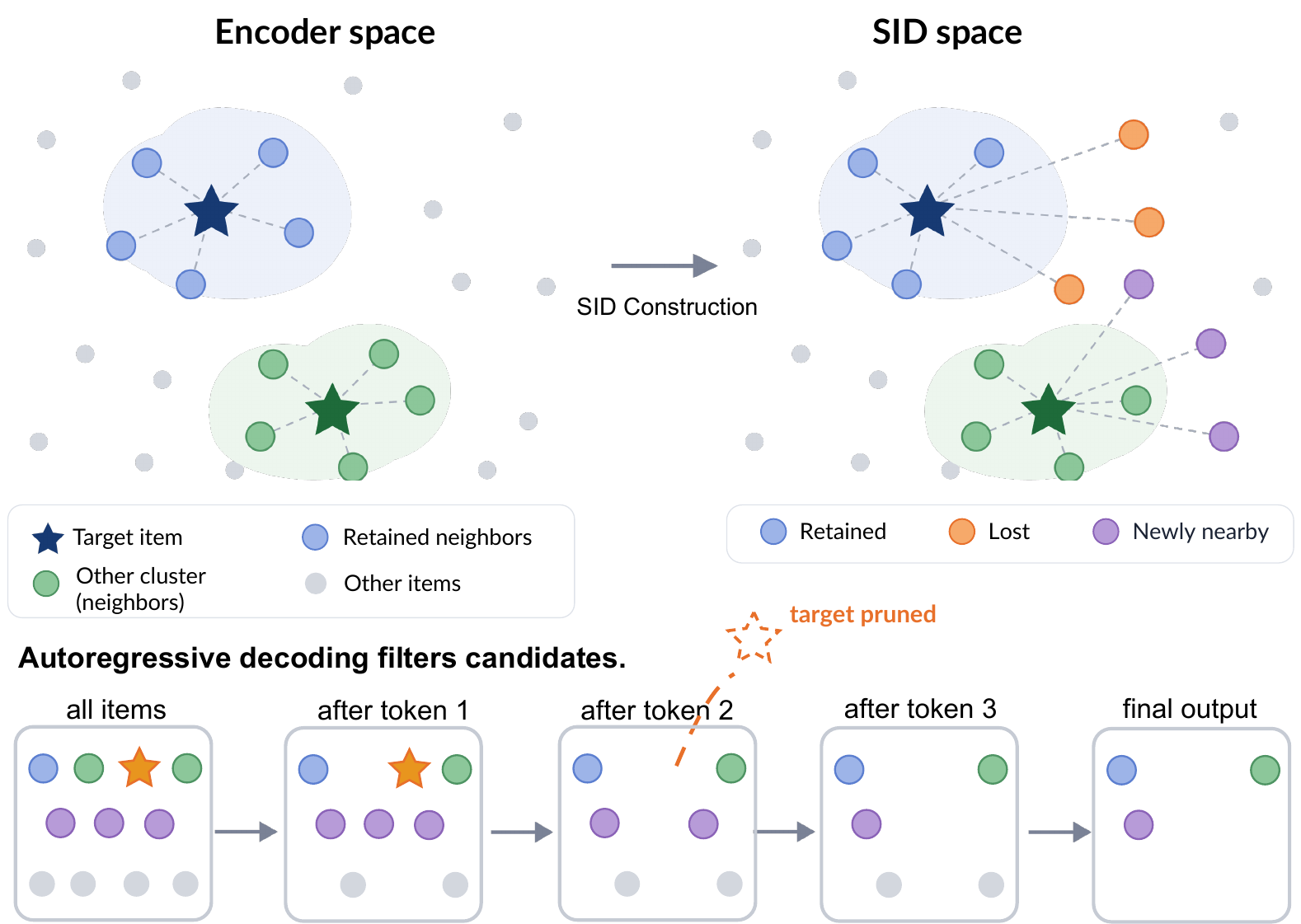 }
  \caption{SIDs preserve broad semantic organization but lose fine-grained local item structure.}
  \Description{A continuous left-to-right diagram summarizes the paper. Two
  descriptions of the same item pass through the same encoder but receive
  different exact SIDs. During generation, the decoder prefers token a even
  though the held-out target lies under token b; choosing token a removes the
  target from all later decisions. ISD constructs a ranked item set from
  training interactions, maps its strongest items to their SID prefixes, and
  combines that support with the decoder rank before beam reduction, keeping the
  target available.}
  \label{fig:intro-story}
  \vspace{-15pt}
\end{figure}

\textbf{We ask two questions.} \textit{What changes when item representations are converted into SIDs? How do the resulting SIDs shape item selection during generation?} We answer the first by testing whether equivalent descriptions produce the same exact SID and whether the organization of the continuous item space survives discretization. We answer the second by tracing how the set of items that can still be recommended changes after each generated token.



To study how item representations change during SID construction, we examine both the content encoder and the SID constructor (\Cref{sec:measurement}). Following the principle of controlled behavioral testing~\cite{ribeiro2020checklist}, for each item, we create an alternative description containing exactly the same words but presenting its fields in a different order, such as placing the title before rather than after the description. Across three encoder families, the original and alternative descriptions have an average cosine similarity of approximately 0.99. In nearest-neighbor retrieval over all original item representations, the alternative representation retrieves its corresponding item first in 99.57\% of the Office and Scientific cases. Across eight SID constructions and three domains, 38.4\% of items receive a different exact SID on average, while the resulting code spaces recover only 32.2\% of the encoder's ten nearest neighbors. Thus, SIDs preserve broad organization but lose much of the encoder's fine structure, and semantic identity alone does not determine their exact tokens.

To study how the resulting SIDs shape item selection, we follow each held-out target through SID generation and record the token at which it can no longer be returned (\Cref{sec:discretization}). Each generated token narrows the remaining item set, and removing an SID prefix permanently removes every item whose SID begins with it. Only 5.1\% of targets remain before final ranking. A separate recommender then ranks all eligible items. With the SIDs and beam size fixed, allowing highly ranked items to support their prefixes raises target availability to 8.5\%.

Together, these results expose a mismatch between how SIDs organize items and how they are used during generation. SID tokens must distinguish individual items, yet beam search also uses those tokens to decide which items remain eligible. A distinction needed for identification can therefore become a premature relevance decision.
  This motivates our method, using SIDs to organize and identify items while introducing separate item-level evidence to determine which items remain under consideration.

\textbf{Present work.} We propose \emph{Item-Supported Decoding} (ISD), a lightweight decoding intervention that requires no additional parameters or retraining. Given an item ranking derived from the user's history, ISD gives additional weight to SID prefixes associated with highly ranked items before beam reduction. A prefix can therefore remain because the decoder favors it or because it contains an item ranked highly for the user. After generation, ISD uses the same ranking to order the generated items.

Recent work reduces dependence on early SID decisions through architectural changes. SETRec predicts an order-agnostic token set instead of an ordered SID~\cite{lin2025order}. LIGER retains SID generation but adds dense retrieval as a second path for missed items~\cite{yang2024liger}. ISD instead operates inside an existing autoregressive decoder before beam reduction makes a token decision irreversible. Across the evaluated systems, ISD improves NDCG@10 by up to 31.2\%. Our \textbf{key contributions} are as follows:

\begin{description}[labelindent=5pt, labelsep=0pt, topsep=0pt, leftmargin=!, font=\normalfont\bfseries]
\item[Controlled Investigation of SID Structure: ] We introduce a \\same-item evaluation that separates continuous item recognition, exact SID assignment, and local-structure preservation. Unlike evaluations based only on reconstruction or recommendation accuracy, our analysis tests whether descriptions containing the same item facts remain equivalent after SID construction. Across three domains and eight SID constructions, we show that current SIDs preserve broad semantic organization but lose much of the encoder's fine local structure, while their exact tokens are not determined by semantic identity alone.

\item[Consequences for Autoregressive Generation: ]  We connect \\ SID structure to generation by tracing held-out targets after every semantic token. Our analysis shows how fine SID distinctions become irreversible item-selection decisions during beam search. A controlled intervention further demonstrates that allowing item-level evidence to act before beam reduction keeps more ground-truth targets available under the same SIDs and beam size.

\item[Item-Supported Decoding: ] Based on these findings, we propose ISD, a lightweight inference-time method that requires no retraining of the SID constructor or decoder. ISD accepts an item ranking from any ranking method, uses it to support relevant SID prefixes before beam reduction, and orders only the generated items afterward. Across TIGER and LIGER, three domains, and rankings produced by training statistics, SASRec, and UniSRec, ISD consistently improves NDCG@10, with gains of up to 31.2\%.
\end{description}

\begin{figure*}[t]
  \centering
  \includegraphics[width=\textwidth]{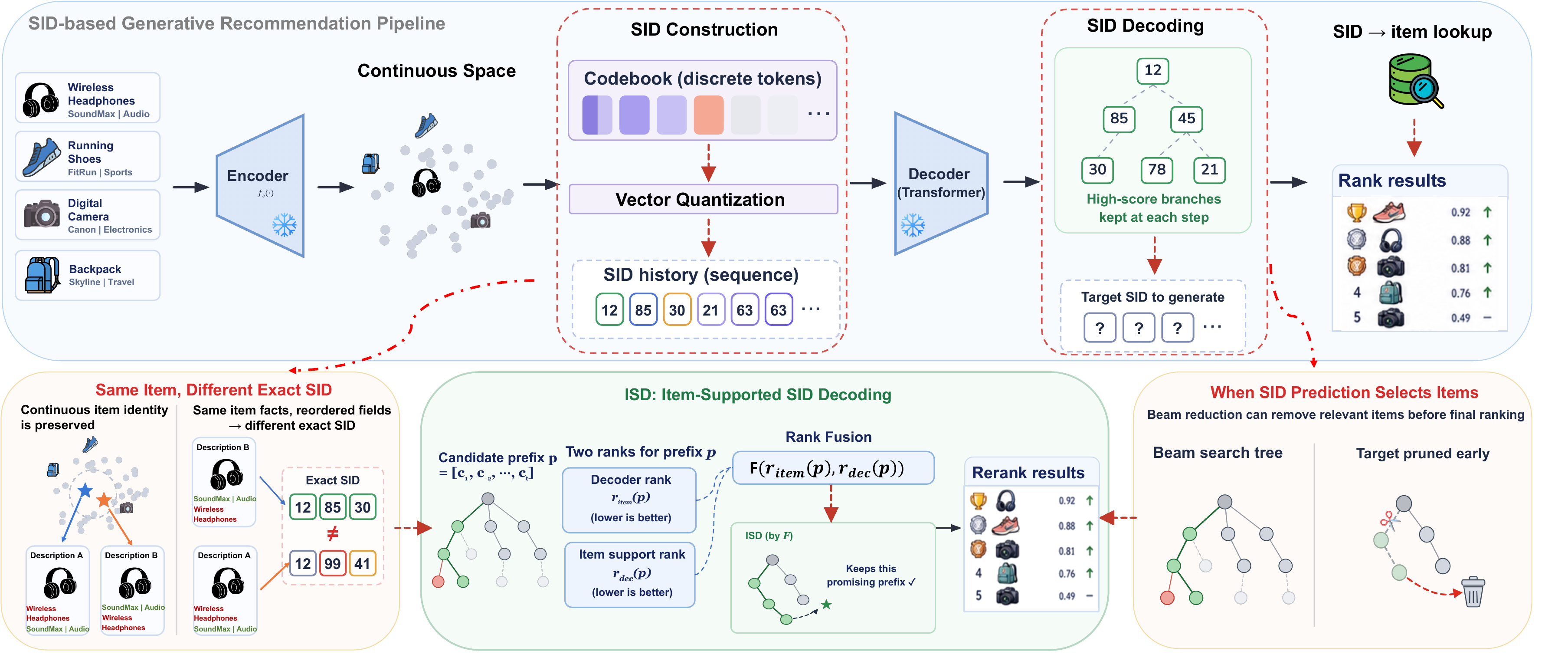}
  \vspace{-10pt}
  \caption{Overview of SID construction, item filtering, and Item-Supported Decoding (ISD).}
  \Description{A continuous left-to-right diagram summarizes the paper. Two
  descriptions of the same item pass through the same encoder but receive
  different exact SIDs. During generation, the decoder prefers token a even
  though the held-out target lies under token b; choosing token a removes the
  target from all later decisions. ISD constructs a ranked item set from
  training interactions, maps its strongest items to their SID prefixes, and
  combines that support with the decoder rank before beam reduction, keeping the
  target available.}
  \label{fig:pipeline-overview}
  \vspace{-10pt}
\end{figure*}

\section{Preliminaries}
\label{sec:preliminaries}
This section defines the generative recommendation task and the notation used throughout the paper.

\paragraph{\textbf{\textit{Generative recommendation.}}} Let $\mathcal I=\{1,\ldots,N\}$ be the item set and $\mathcal U$ the user set. For a user $u\in\mathcal U$, the interaction history $H_u=(i_1,\ldots,i_T)$ is an ordered sequence of items from $\mathcal I$. Given $H_u$, the next-item recommendation task is to rank a target item $i^+\in\mathcal I$ ahead of the remaining items. We write $\mathcal D=\{(H_u,i^+)\}$ for the training examples.

\paragraph{\textbf{\textit{Semantic IDs.}}} Each item $i$ is associated with content $x_i$, such as its title, brand, category, or description. An SID pipeline encodes this content as a continuous representation $h_i=f(x_i)$ and discretizes it into a semantic tuple $z_i=q(h_i)=(z_{i,1},\ldots,z_{i,L})$~\cite{lee2022rqvae,rajput2023recommender,wang2024letter}. Some constructors additionally apply a collision-resolution or uniqueness rule $a$~\cite{rajput2023recommender}; we therefore write
\begin{equation}
  h_i=f(x_i),\qquad z_i=q(h_i),\qquad C_i=a(z_i,i),
  \label{eq:sid-pipeline}
\end{equation}
where $C_i=(c_{i,1},\ldots,c_{i,L_i})$ is the complete SID used by the recommender. When no additional rule is required, $a$ is the identity map and $C_i=z_i$. This notation separates the continuous item representation $h_i$, the content-derived discrete tuple $z_i$, and the complete item sequence $C_i$ presented to the recommendation model.

\paragraph{\textbf{\textit{SID-based prediction.}}} Let $C(H_u)=(C_{i_1},\ldots,C_{i_T})$ be the SID representation of a user's history. A generative recommender assigns probability $p_\theta(C_i\mid C(H_u))$ to an item's complete SID and is trained by~\cite{rajput2023recommender}
\begin{equation}
  \mathcal L_{\mathrm{rec}}(\theta)=-\sum_{(H_u,i^+)\in\mathcal D}\log p_\theta\!\left(C_{i^+}\mid C(H_u)\right).
  \label{eq:generative-rec-objective}
\end{equation}
For an autoregressive decoder, this sequence probability factorizes as $p_\theta(C_i\mid C(H_u))=\prod_{\ell=1}^{L_i}p_\theta(c_{i,\ell}\mid C(H_u),c_{i,<\ell})$~\cite{rajput2023recommender}; parallel decoders use a different factorization but predict the same complete item sequence~\cite{hou2025rpg}. At inference, generated SIDs are resolved to items through the system's SID-to-item map, and items are ranked by their sequence scores. Later sections distinguish the semantic tuple from the complete SID whenever a constructor or decoding rule makes that distinction necessary.


\section{What Semantic IDs Preserve and Lose}
\label{sec:measurement}

Semantic IDs rest on a common premise: item content can provide both the structure used to share information across related items and the discrete label used to identify an item during generation~\cite{rajput2023recommender,wang2024letter,liu2024etegrec,hou2025rpg,fu2026diger}. SIDs are expected to put similar items under similar codes while assigning an identifiable sequence to each item. We study what changes during this conversion from two perspectives. First, we test whether equivalent descriptions of the same item produce the same exact SID. Second, we test how much of the continuous item-space organization remains after SID construction. These properties matter because the exact SID is the supervised prediction target, while shared structure is intended to let related items share statistical strength.

We trace items through content encoding
(\Cref{sec:exact-sid-assignment}), SID construction, and decoding. The
constructor comparison is reported in \Cref{tab:primary-multiseed}, while
\Cref{tab:encoder-encoder-neighborhood} provides the continuous encoder
controls.

\subsection{Exact SID Assignment}
\label{sec:exact-sid-assignment}
This section tests what determines an item's exact SID. We test what happens when the same item facts are presented in different orders. If an SID depends only on the item's semantic meaning, its assigned code should remain unchanged.

\subsubsection{\textbf{Analysis Setup}}
Following behavioral invariance testing~\cite{ribeiro2020checklist}, we change presentation while preserving the item and its contents. Let $\mathcal X$ denote the set of item descriptions and let $\pi(x)\in\mathcal I$ be the item described by $x$. Starting from source description $x_i$, we construct $x_i'=\tau(x_i)$ by changing only the order in which the same item information, such as its title and description, is presented. No words or attribute values are added, removed, or changed. We write $x\sim x'$ when two descriptions refer to the same item, so that $\pi(x)=\pi(x')$. Thus, every evaluated pair satisfies
\begin{equation}
  x_i\sim x_i',\qquad \operatorname{bag}(x_i)=\operatorname{bag}(x_i'),
  \qquad \pi(x_i)=\pi(x_i')=i,
  \label{eq:matched-description}
\end{equation}
where $\operatorname{bag}(x)$ is the multiset of tokens and field values in $x$. Thus $x_i$ and $x_i'$ differ only in presentation. Let $\mathcal P$ be the set of item indices represented by these matched-description pairs. If the complete SID $C(x)$ identifies the item directly, it must be constant over such matched descriptions:
\begin{equation}
  x\sim x' \quad\Longrightarrow\quad C(x)=C(x').
  \label{eq:item-label}
\end{equation}    
\begin{table}[t]
  \centering
  \caption{Encoder stability (\%).}
  \vspace{-5pt}
  \label{tab:encoder-encoder-neighborhood}
  \small
  \setlength{\tabcolsep}{4.0pt}
  \begin{tabular}{@{}lcccc@{}}
    \toprule
    & \multicolumn{2}{c}{Office} & \multicolumn{2}{c}{Scientific} \\
    \cmidrule(lr){2-3}\cmidrule(l){4-5}
    Text encoder & Identity@1 & $R_{\mathrm{Enc}}^{10}$ & Identity@1 & $R_{\mathrm{Enc}}^{10}$ \\
    \midrule
    Sentence-T5 & 99.51 & 91.81 & 99.51 & 92.72 \\
    BGE-base & 99.46 & 91.55 & 99.46 & 92.87 \\
    Qwen3-0.6B & 98.97 & 86.10 & 99.41 & 89.96 \\
    \bottomrule
  \end{tabular}
  \vspace{-10pt}
\end{table}

\subsubsection{\textbf{Metric and Interpretation}}

For each $i\in\mathcal P$, let $C(x_i)$ and $C(x_i')$ be the complete SIDs produced by the same frozen constructor. We measure
\begin{equation}
  M_i=\mathbf{1}\!\left[C(x_i)\neq C(x_i')\right],
  \qquad
  \Delta_{\mathrm{SID}}=\frac{1}{|\mathcal P|}\sum_{i\in\mathcal P}M_i.
  \label{eq:exact-change}
\end{equation}
Intuitively, $M_i=1$ means that this order change gives item $i$ a different complete SID, \ie the token sequence that the recommender is trained to generate. $\Delta_{\mathrm{SID}}$ is the proportion of matched items for which this occurs.
We present the results in~\Cref{tab:primary-multiseed}. The RQ-VAE-latent
control applies RQ-KMeans to the representation learned by RQ-VAE, changing
the quantizer while holding its input fixed.
\begin{table*}[t]
  \centering
  \caption{Exact-code change and top-10 neighborhood preservation, using Sentence-T5 as the encoder}
  \label{tab:primary-multiseed}
  \label{tab:primary-neighborhood}
  \small
  \setlength{\tabcolsep}{1.0pt}
  \begin{tabular}{@{}lccc@{\hspace{5pt}}ccc@{\hspace{5pt}}ccc@{}}
    \toprule
    & \multicolumn{3}{c}{Baby (\%)} & \multicolumn{3}{c}{Office (\%)} & \multicolumn{3}{c}{Scientific (\%)} \\
    \cmidrule(lr){2-4}\cmidrule(lr){5-7}\cmidrule(l){8-10}
    SID constructor & $\Delta$SID$\downarrow$ & Enc--SID $F^{10}\uparrow$ & SID--SID$'$ $R^{10}_{\text{SID}}\uparrow$ & $\Delta$SID$\downarrow$ & Enc--SID $F^{10}\uparrow$ & SID--SID$'$ $R^{10}_{\text{SID}}\uparrow$ & $\Delta$SID$\downarrow$ & Enc--SID $F^{10}\uparrow$ & SID--SID$'$ $R^{10}_{\text{SID}}\uparrow$ \\
    \midrule
    RQ-VAE~\cite{lee2022rqvae,rajput2023recommender} & 54.5 & 19.9 & 71.1 & 53.9 & 13.8 & 63.2 & 45.7 & 32.8 & 80.4 \\
    RQ-KMeans~\cite{penha2025semanticids} & 20.4 & 36.5 & 94.6 & 29.3 & 31.5 & 87.0 & 17.5 & 43.2 & 96.5 \\
    RQ-KMeans (normalized)~\cite{penha2025semanticids} & \textbf{17.4} & \textbf{38.9} & \textbf{96.0} & 24.6 & \textbf{33.1} & \textbf{89.0} & \textbf{14.0} & \textbf{45.1} & \textbf{97.6} \\
    RQ-KMeans (RQ-VAE latent)~\cite{penha2025semanticids,lee2022rqvae} & 50.8 & 23.8 & 79.5 & 63.8 & 18.7 & 66.9 & 41.2 & 37.5 & 87.5 \\
    DRQ~\cite{decoupledrq2026} & 35.4 & 33.9 & 85.2 & 42.3 & 27.9 & 76.4 & 29.5 & 44.3 & 90.8 \\
    RQP-VAE~\cite{decoupledrq2026} & 73.5 & 36.6 & 79.0 & 58.0 & 32.1 & 73.2 & 67.0 & 45.0 & 84.2 \\
    OPQ/PQ~\cite{ge2013opq,hou2025rpg} & 26.3 & 26.1 & 86.7 & \textbf{20.7} & 27.7 & 88.4 & 29.4 & 18.3 & 80.7 \\
    Balanced assignment~\cite{fu2026diger} & 33.2 & 36.6 & 87.7 & 36.2 & 28.2 & 82.5 & 38.1 & 40.6 & 88.9 \\
    \bottomrule
  \end{tabular}
  \vspace{-10pt}
\end{table*}

\subsubsection{\textbf{Encoder-Space Control}}
Before measuring SID changes, we verify that reordering preserves the item's identity in the continuous embedding space. Let $e_i=f(x_i)$ and $e_i'=f(x_i')$ denote the representations of item $i$ under its original and reordered descriptions. We measure their cosine similarity and use $e_i'$ to query the original item representations $\{e_j=f(x_j)\}_{j\in\mathcal{I}}$. We define identity-retrieval accuracy as
\begin{equation}
\operatorname{Identity@1}
=
\frac{1}{|\mathcal{I}|}
\sum_{i\in\mathcal{I}}
\mathbb{I}\!\left[
i=\arg\max_{j\in\mathcal{I}}\cos(e_i',e_j)
\right].
\end{equation}
The reordered representation $e_i'$ is not included in the item index. Under Sentence-T5, the mean cosine similarity between $e_i$ and $e_i'$ is $0.9939$. Across Sentence-T5, BGE, and Qwen, the reordered description retrieves the original item first in 98.97--99.51\% of fixed-anchor cases and retains 86.10--92.87\% of its original top-10 neighbors. Thus, the encoders continue to recognize the same item and largely preserve its local neighborhood. Table~\ref{tab:encoder-encoder-neighborhood} reports these controls.


\subsubsection{\textbf{Exact SIDs Are Not Determined by Meaning Alone}}

Reordering the same item facts preserves continuous item recognition, \ie the reordered description retrieves the original item first in 99.57\% of controlled cases. However, 14.0--73.5\% of items receive a different SID across constructors and domains (\Cref{tab:primary-multiseed}). Thus, recognizing the same item does not guarantee the same discrete code. Since this code is used as the generation target, the decoder must learn the constructor's chosen assignment in addition to the item relationships encoded by the SID.

\subsection{Local-Structure Preservation}
\label{sec:local-structure}
In this section, we measure how faithfully SIDs preserve the local structure of the continuous item space from which they are built.
\subsubsection{\textbf{Metrics and Interpretation}}


For item $i$, let $N_E^k(x_i)$ be its self-excluded top-$k$ encoder neighbors and $N_S^k(x_i)$ its top-$k$ neighbors after SID reconstruction. We define
\begin{equation}
  F_i^k=\frac{|N_E^k(x_i)\cap N_S^k(x_i)|}{k},
  \qquad
  R_{\mathrm{SID},i}^k=\frac{|N_S^k(x_i)\cap N_S^k(x_i')|}{k}.
  \label{eq:neighborhood-metrics}
\end{equation}
We report their averages over the evaluated items as $F^k$ and $R_{\mathrm{SID}}^k$. $F_i^k$ measures encoder-neighbor recovery for item $i$, while $R_{\mathrm{SID},i}^k$ measures whether its SID neighbors survive the field reorder. The corresponding encoder-only control, $R_{\mathrm{Enc}}^k=|N_E^k(x_i)\cap N_E^k(x_i')|/k$, appears in Table~\ref{tab:encoder-encoder-neighborhood}.



\subsubsection{\textbf{SIDs Preserve Broad Organization, Not Fine Neighborhoods}}

Table~\ref{tab:primary-neighborhood} shows that even before the order test, SID reconstructions recover only 13.8--45.1\% of the encoder's exact top-10 neighbors. Additional ranking and distance checks show that missing exact neighbors are usually replaced by other nearby items (\Cref{app:neighborhood}). The SIDs therefore retain broad organization while substantially rewriting the fine local structure from which they were built. The same distinction appears under the order test: SID neighborhoods retain 63.2--97.6\% of their members, compared with 91.8--92.7\% in the continuous-space control. \Cref{app:sid-geometry-visualizations} provides the corresponding geometry visualizations.

\begin{figure}[h]
  \centering
  \includegraphics[width=\linewidth]{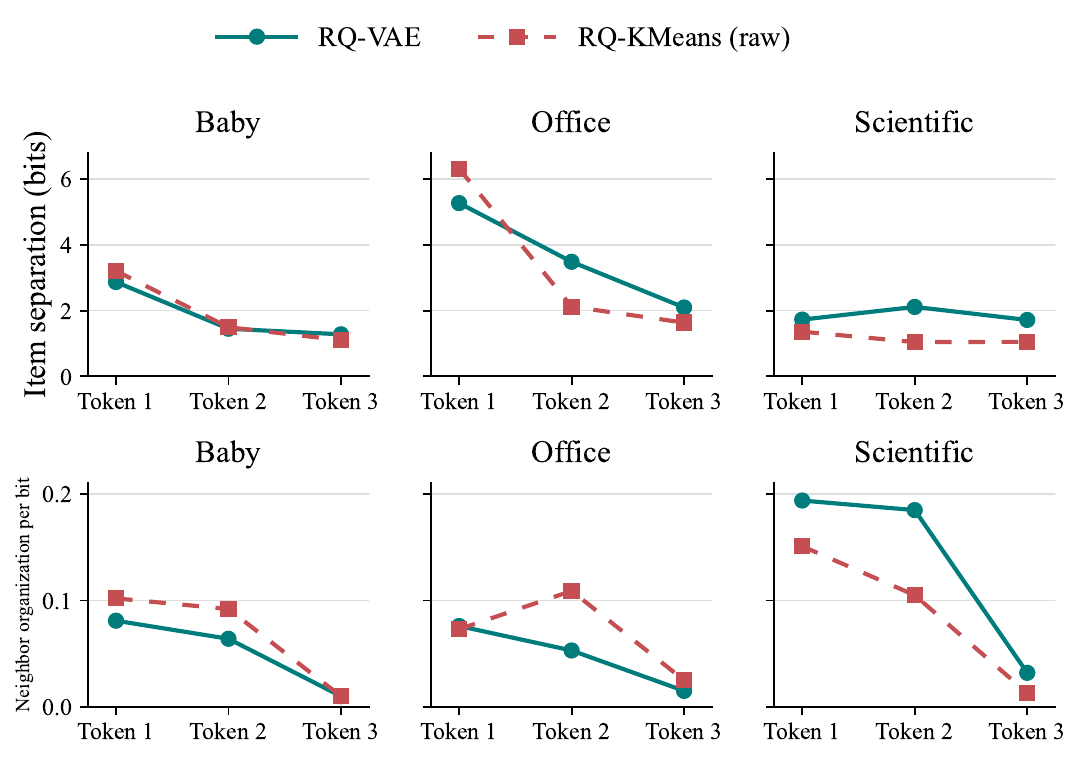}
  \caption{Item separation and source-neighbor organization across the three semantic tokens.}
  \label{fig:token-roles}
  \vspace{-15pt}
\end{figure}

\subsection{Coarse Organization and Fine Identification}
The analyses in \Cref{sec:exact-sid-assignment,sec:local-structure} show that complete SIDs preserve broad organization but rewrite many fine item relationships. We next measure how item separation and neighbor preservation change across the SID. Following~\cite{rajput2023recommender,fu2026diger}, we set the token length to three and measure at each semantic position how much the next token reduces the possible items and whether encoder-space neighbors remain together. We exclude the collision-resolution suffix. \Cref{app:position-organization} defines the measurements and randomization protocol.

Across both constructors and all three datasets, the first two tokens reduce the possible items while keeping substantially more encoder-space neighbors together. The final token continues to reduce the possible items, but keeps few additional neighbors together beyond the randomized expectation (\Cref{fig:token-roles}). It therefore contributes mainly to distinguishing items within groups formed by the earlier tokens.

Separating items is necessary to identify the final recommendation. The concern is that autoregressive decoding applies this separation before the complete SID is known. At each position, beam search discards some SID prefixes and removes every item whose identifier begins with a discarded prefix. \Cref{sec:fixed-map-attribution} measures how many otherwise plausible targets are lost during this process.

\section{When SID Prediction Selects Items}
\label{sec:discretization}
\label{sec:fixed-map-attribution}

\Cref{sec:measurement} shows that early SID tokens place many encoder-space
neighbors under the same prefix, while later tokens increasingly separate
individual items. Autoregressive generation treats every token as a selection
step. After predicting each token, beam search retains a fixed number of SID
prefixes and discards the rest. Discarding a prefix removes every item whose
SID begins with it, even though the final item has not yet been selected.
A low score on a later token can therefore remove the correct item before its full SID is generated or final ranking begins.

The SID constructor determines which items share each prefix. When two
encoder-space neighbors receive different later tokens, the decoder must
assign probability to two different token sequences rather than one shared
prefix.
Although the decoder receives the user's history, beam search ranks each SID
prefix as a whole rather than ranking the individual items assigned to it. If the
target's prefix is discarded, the target is removed even when a separate model
would rank that item highly from the same history. \Cref{sec:target_rank}
measures how often this occurs.


Prior work has addressed early removal by changing the recommendation architecture. SETRec predicts an unordered token set instead of an ordered SID~\cite{lin2025order}. LIGER combines SID generation with dense retrieval and ranking~\cite{yang2024liger}. We instead keep the SID assignment, decoder, and beam search fixed.
We follow each held-out target through SID generation and record when its
prefix leaves the beam. Section~\ref{sec:target_rank} identifies removed
targets that a separate item model ranks among its top $k$ items.
Section~\ref{sec:target_rank_intervention} then tests whether using these item
scores during beam selection keeps more targets available.

\subsection{How SID Prediction Filters Items}

Suppose the decoder has generated SID prefix $p$, the sequence of SID tokens produced so far. Let $S_p$ be the set of items whose assigned SIDs begin with $p$. For a possible next token $v$, the smaller set $S_{p,v}\subseteq S_p$ contains the items whose SIDs begin with $(p,v)$. We write $\mathcal V(p)=\{v:S_{p,v}\neq\varnothing\}$ for the possible next tokens. The decoder scores each $v\in\mathcal V(p)$ using the user's interaction history $h$ and the complete prefix,
\begin{equation}
  p_\theta(v\mid h,p).
  \label{eq:decoder-next-token}
\end{equation}
Beam search expands each retained SID prefix with its possible next tokens. Its
\emph{beam width} $B$ is the number of extended prefixes retained after each
token~\cite{rajput2023recommender,lin2025order}. If $\mathcal B_\ell$ is the
set of these $B$ prefixes after semantic position $\ell$, then the items still
available for recommendation are
\begin{equation}
  \mathcal R_\ell=\bigcup_{p\in\mathcal B_\ell}S_p,
  \label{eq:reachable-items}
\end{equation}
Removing an SID prefix removes every item in its set $S_p$. If the held-out
target is not in $\mathcal R_\ell$, no later token or final ranking step can
return it. The constructor therefore determines which items share each SID
prefix. Beam search then uses those prefixes to decide which items remain
available.

\subsection{Are Highly Ranked Targets Removed Early?}
\label{sec:target_rank}
SID generation removes some held-out targets before final item ranking begins.
A removal affects ranking at cutoff $k$ only when the target could otherwise
appear among the top $k$ items. We use a separate model that ranks items
directly from the user's history to identify these targets, then measure how
many remain after each SID token.

\subsubsection{\textbf{Analysis Setup}}
We train a separate sequential item scorer $g(h,i)$ on training interactions.
Given user history $h$, it ranks every unconsumed item $i$ before SID
generation removes any items. We retain cases in which this scorer places the
held-out target $i^\star$ among its top $k$ items. We then ask whether SID
generation removes these highly ranked targets before final ranking. In the
reported experiment, we set $k=10$ to match Recall@10 and NDCG@10.


\subsubsection{\textbf{Metrics and Interpretation}}
Let $\mathcal H_k$ contain these top-$k$ cases, and let $i^\star(h)$ be the held-out target for history $h$. At each semantic position $\ell$, we measure the fraction that remains available:
\begin{equation}
K_\ell
  =\frac{1}{|\mathcal H_k|}
    \sum_{h\in\mathcal H_k}
    \mathbf 1[i^\star(h)\in\mathcal R_\ell].
\label{eq:rankable-target-retention}
\end{equation}
Thus, $K_\ell$ is the fraction of these top-$k$ targets that can still be
returned after token $\ell$. Appendix~\ref{app:rankable-target-protocol}
reports an additional analysis that gives more weight to targets ranked closer
to first.


\begin{figure}[t]
  \centering
  \includegraphics[width=\columnwidth]{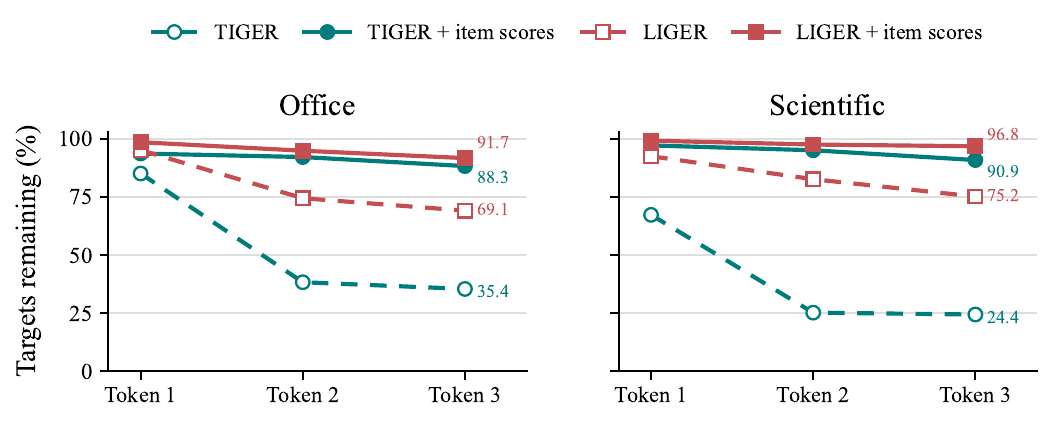}
  \Description{Two line charts show the percentage of top-k held-out targets
  remaining after each semantic token for TIGER and LIGER on Office and
  Scientific, with and without item scores. The reported cutoff is k equals
  ten.}
  
  \caption{Top-$k$ targets remaining after each semantic token, with $k=10$.
  Adding the direct item score uses $\alpha=0.75$. SID assignments and beam
  width are unchanged.}
  \label{fig:rankable-target-filtering}
  \vspace{-10pt}
\end{figure}

\subsubsection{\textbf{Removal Precedes Final Ranking}}
\Cref{fig:rankable-target-filtering} shows that both decoders remove top-$k$
targets during SID generation, although TIGER removes more. With $k=10$, after
the third token TIGER retains 35.4\% of the Office targets and 24.4\% of the
Scientific targets.
LIGER retains 69.1\% and 75.2\%, respectively. Thus, SID prediction can remove
targets that the separate scorer would place near the top before final item
ranking begins.

\subsection{Item Evidence Before Filtering}
\label{sec:target_rank_intervention}
\Cref{fig:rankable-target-filtering} shows that SID generation removes many
targets that the separate item scorer ranks highly. We next test whether these
removals can be reduced without changing the SID assignments or retraining the
decoder. The intervention increases the score of an extended prefix when it
contains an item ranked highly by the separate scorer.

\subsubsection{\textbf{Intervention Before Prefix Removal}}
At a decoding step, prefix $p$ can be extended with any token $v$ for which
$S_{p,v}$ is nonempty. Ordinary beam search scores each extension $(p,v)$
using the decoder probability. We additionally take the highest score among
the unconsumed items that the extension would keep:
\begin{equation}
  q(h,p,v)=\max_{i\in S_{p,v}\setminus h}g(h,i).
  \label{eq:branch-lookahead}
\end{equation}
A large $q(h,p,v)$ means that choosing token $v$ keeps at least one item that
the separate scorer ranks highly. The held-out target identity is never
supplied. Let $d_v=\log p_\theta(v\mid h,p)$ be the decoder score. We combine
the two scores as
\begin{equation}
  s_\alpha(h,p,v)=(1-\alpha)\widetilde d_v+\alpha\widetilde q_v, 
  \label{eq:lookahead-combination}
\end{equation}
where $\widetilde d_v$ and $\widetilde q_v$ denote the normalized decoder and
item scores across the possible next tokens.
\subsubsection{\textbf{Item Scores Preserve More Targets}}
\Cref{fig:rankable-target-filtering} shows that using a separate item scorer to
guide beam search substantially increases the number of targets retained.
However, the intervention is expensive because it scores every item at every
token position. This motivates the efficient repair in
\Cref{sec:repair-discussion}, which combines the decoder score with one
additional item ranking.

\FloatBarrier

\section{Item-Supported Decoding}
\label{sec:repair-discussion}
Sections~\ref{sec:measurement} and~\ref{sec:discretization} expose a conflict between the roles assigned to an SID. Shared prefixes provide a compact organization of the item set, but the fine tokens do not preserve every fine-grained item relationship. During generation, those same tokens become filtering decisions: removing an SID prefix removes all items assigned beneath it. Section~\ref{sec:discretization} shows that many targets remain available when token selection also considers whether each token would preserve an item that a separate sequential recommender ranks highly from the user's history. That diagnostic repeatedly scores every unconsumed item at every token position, making it too expensive to use as a recommender.

We turn that diagnostic into a lightweight repair called \emph{Item-Supported Decoding} (ISD). Given the observed history, ISD constructs one ranked item set from a separate item ranker (\Cref{sec:ranking_set}). During SID generation, this ranking supports prefixes before beam reduction removes their items (\Cref{sec:before-filtering}). After complete SIDs are generated, the same ranking orders the generated items (\Cref{sec:after-completion}).

\begin{table*}[!t]
  \centering
  \caption{Recommendation results. Each \% Improvement row compares the ISD variant immediately above it with the corresponding generative recommender. Bold and underlined values indicate the best and second-best results in each column. N@K and R@K denote NDCG and Recall at cutoff $K$.}
  \vspace{-5pt}
  \label{tab:recommendation-results}
  \label{tab:evidence-fusion-results}
  \small
  \setlength{\tabcolsep}{0.2pt}
  \renewcommand{\arraystretch}{0.90}
  \begin{tabular*}{\textwidth}{@{\extracolsep{\fill}}lrrrrrrrrrrrr@{}}
    \toprule
    & \multicolumn{4}{c}{Scientific} & \multicolumn{4}{c}{Office} & \multicolumn{4}{c}{Baby} \\
    \cmidrule(lr){2-5}\cmidrule(lr){6-9}\cmidrule(l){10-13}
    Method
      & N@10 & N@20 & R@10 & R@20
      & N@10 & N@20 & R@10 & R@20
      & N@10 & N@20 & R@10 & R@20 \\
    \midrule
    SASRec~\cite{kang2018sasrec}
      & 0.0844 & 0.0926 & 0.1183 & 0.1509
      & 0.1008 & 0.1062 & 0.1282 & 0.1495
      & 0.0347 & 0.0420 & 0.0623 & 0.0912 \\
    BERT4Rec~\cite{sun2019bert4rec}
      & 0.0679 & 0.0745 & 0.0952 & 0.1214
      & 0.0909 & 0.0951 & 0.1143 & 0.1313
      & 0.0185 & 0.0240 & 0.0359 & 0.0580 \\
    UniSRec~\cite{hou2022unisrec}
      & 0.0842 & 0.0910 & 0.1169 & 0.1440
      & 0.1046 & 0.1102 & 0.1342 & 0.1564
      & 0.0296 & 0.0376 & 0.0653 & 0.0972 \\
    SETRec~\cite{lin2025order}
      & 0.0822 & 0.0904 & 0.1190 & 0.1518
      & 0.1032 & 0.1086 & 0.1313 & 0.1530
      & 0.0353 & 0.0426 & 0.0652 & 0.0942 \\
    ETEGRec~\cite{liu2024etegrec}
      & 0.0783 & 0.0860 & 0.1108 & 0.1414
      & 0.1024 & 0.1075 & 0.1298 & 0.1486
      & 0.0337 & 0.0412 & 0.0637 & 0.0933 \\
    Statistics ranker~\cite{rendle2010fpmc,sarwar2001item}
      & 0.0824 & 0.0882 & 0.1098 & 0.1331
      & 0.1016 & 0.1049 & 0.1247 & 0.1381
      & 0.0299 & 0.0337 & 0.0479 & 0.0633 \\
    \midrule
    TIGER~\cite{rajput2023recommender}
      & 0.0802 & 0.0887 & 0.1167 & 0.1506
      & 0.0949 & 0.1005 & 0.1194 & 0.1415
      & 0.0295 & 0.0371 & 0.0565 & 0.0866 \\
    \quad + ISD (statistics)
      & 0.0891 & 0.0947 & 0.1235 & 0.1460
      & \underline{0.1074} & \underline{0.1119} & 0.1349 & 0.1527
      & 0.0349 & 0.0391 & 0.0579 & 0.0745 \\
    \textit{\% Improvement}
      & \textit{+11.0\%} & \textit{+6.8\%} & \textit{+5.8\%} & \textit{$-$3.1\%}
      & \textit{+13.1\%} & \textit{+11.3\%} & \textit{+13.0\%} & \textit{+7.9\%}
      & \textit{+18.2\%} & \textit{+5.5\%} & \textit{+2.5\%} & \textit{$-$14.0\%} \\
    \quad + ISD (SASRec)
      & 0.0875 & 0.0957 & 0.1277 & 0.1604
      & 0.1026 & 0.1081 & 0.1332 & 0.1549
      & 0.0365 & 0.0444 & 0.0686 & 0.0997 \\
    \textit{\% Improvement}
      & \textit{+9.1\%} & \textit{+7.9\%} & \textit{+9.5\%} & \textit{+6.5\%}
      & \textit{+8.1\%} & \textit{+7.5\%} & \textit{+11.5\%} & \textit{+9.5\%}
      & \textit{+23.9\%} & \textit{+19.7\%} & \textit{+21.7\%} & \textit{+15.1\%} \\
    \quad + ISD (UniSRec)
      & 0.0885 & 0.0965 & 0.1257 & 0.1572
      & \underline{0.1048} & \underline{0.1109} & \textbf{0.1367} & \textbf{0.1606}
      & 0.0387 & 0.0482 & 0.0716 & \underline{0.1093} \\
    \textit{\% Improvement}
      & \textit{+10.3\%} & \textit{+8.7\%} & \textit{+7.8\%} & \textit{+4.4\%}
      & \textit{+10.4\%} & \textit{+10.3\%} & \textit{+14.5\%} & \textit{+13.5\%}
      & \textit{+31.2\%} & \textit{+30.0\%} & \textit{+27.1\%} & \textit{+26.2\%} \\
    \addlinespace
    LIGER~\cite{yang2024liger}
      & 0.0880 & 0.0970 & 0.1291 & 0.1650
      & 0.0983 & 0.1038 & 0.1257 & 0.1476
      & 0.0398 & 0.0480 & 0.0726 & 0.1053 \\
    \quad + ISD (statistics)
      & \textbf{0.0927} & 0.0986 & 0.1266 & 0.1499
      & \textbf{0.1081} & \textbf{0.1126} & \underline{0.1357} & 0.1534
      & 0.0403 & 0.0483 & 0.0734 & 0.1052 \\
    \textit{\% Improvement}
      & \textit{+5.3\%} & \textit{+1.6\%} & \textit{$-$1.9\%} & \textit{$-$9.1\%}
      & \textit{+10.1\%} & \textit{+8.5\%} & \textit{+8.0\%} & \textit{+3.9\%}
      & \textit{+1.1\%} & \textit{+0.5\%} & \textit{+1.2\%} & \textit{$-$0.1\%} \\
    \quad + ISD (SASRec)
      & 0.0914 & \underline{0.1010} & \underline{0.1335} & \underline{0.1714}
      & 0.1010 & 0.1075 & 0.1314 & 0.1571
      & \underline{0.0406} & \underline{0.0491} & \underline{0.0745} & 0.1082 \\
    \textit{\% Improvement}
      & \textit{+3.8\%} & \textit{+4.1\%} & \textit{+3.4\%} & \textit{+3.9\%}
      & \textit{+2.8\%} & \textit{+3.6\%} & \textit{+4.6\%} & \textit{+6.4\%}
      & \textit{+1.8\%} & \textit{+2.0\%} & \textit{+2.5\%} & \textit{+2.6\%} \\
    \quad + ISD (UniSRec)
      & \underline{0.0921} & \textbf{0.1013} & \textbf{0.1356} & \textbf{0.1724}
      & 0.1011 & 0.1081 & 0.1317 & \underline{0.1593}
      & \textbf{0.0416} & \textbf{0.0509} & \textbf{0.0769} & \textbf{0.1137} \\
    \textit{\% Improvement}
      & \textit{+4.6\%} & \textit{+4.4\%} & \textit{+5.1\%} & \textit{+4.5\%}
      & \textit{+2.9\%} & \textit{+4.1\%} & \textit{+4.8\%} & \textit{+7.9\%}
      & \textit{+4.5\%} & \textit{+5.8\%} & \textit{+5.9\%} & \textit{+7.8\%} \\
    \bottomrule
  \end{tabular*}

\end{table*}

\subsection{Constructing the Ranked Item Set}
\label{sec:ranking_set}
Given the observed history $h$, an item ranker assigns a score $s(h,i)$ to every unconsumed item $i$. We sort the items by this score and retain the top $J$ as the ranked item set $Q(h)$. The value $J$ controls how many items can provide support during SID generation. We construct $Q(h)$ once before generation and use item ranks rather than raw scores. ISD can therefore accept rankings from models with different score scales without model-specific calibration~\cite{cormack2009rrf}. We study two sources of item rankings. The first uses statistics computed from training interactions, while the second uses learned sequential recommenders.
\subsubsection{\textbf{Training-Statistics-Based Ranking}}
\label{sec:training-statistics}

Our simplest ranking combines first-order item transitions and cosine-normalized item co-consumption computed from the training sequences~\cite{rendle2010fpmc,sarwar2001item}. Let $N_i$ be the number of occurrences of item $i$, let $T_{j,i}$ count how often $i$ immediately follows $j$, and let $W_{x,i}$ count how often $x$ and $i$ occur within $w$ positions of one another. We define
\begin{equation}
  A_{j,i}=\log(1+T_{j,i}),
  \qquad
  C_{x,i}=\frac{W_{x,i}}{\sqrt{N_xN_i}}.
  \label{eq:training-item-evidence}
\end{equation}
$A_{j,i}$ measures direct transitions. The logarithm prevents a few frequent transitions from dominating the score. $C_{x,i}$ measures nearby consumption, normalized so that two individually popular items do not receive a high score merely because both appear often. For history $h$, let $j$ be its final observed item and let $H_m(h)$ contain its $m$ most recent items. We score an unconsumed item $i$ by
\begin{equation}
  s_{\mathrm{stat}}(h,i)=\frac{1}{2}\,z_h\!\left(A_{j,i}\right)
  +\frac{1}{2}\,z_h\!\left(
  \frac{1}{|H_m(h)|}\sum_{x\in H_m(h)} C_{x,i}\right),
  \label{eq:behavior-evidence}
\end{equation}
Here, $z_h$ standardizes each signal across the unconsumed items for history $h$. Appendix~\ref{app:isd-details} gives the exact rule. Ranking items by $s_{\mathrm{stat}}(h,i)$ produces $Q(h)$. The statistics and scoring rule are fixed using training data. At evaluation, the score uses the observed input history but never the held-out next item.

\subsubsection{\textbf{Sequential-Model-Based Ranking}}
\label{sec:sequential-models}
A trained sequential recommender provides the same interface. SASRec~\cite{kang2018sasrec} scores items from learned item IDs, while UniSRec~\cite{hou2022unisrec} also incorporates item content. For either model, $s(h,i)$ is its score for unconsumed item $i$. The three reported variants are therefore ISD (statistics), ISD (SASRec), and ISD (UniSRec). The SID model and fusion rule remain the same across ranking sources.
\subsection{Using Item Ranks During Decoding}
\label{sec:isd-decoding}
ISD uses the ranked item set $Q(h)$ before filtering and after SID completion. Both steps combine the decoder rank with the item rank using reciprocal rank fusion~\cite{cormack2009rrf}.
\subsubsection{\textbf{Before Filtering}}
\label{sec:before-filtering} 
Section~\ref{sec:discretization} tests every item beneath each possible next token. ISD approximates this test with $Q(h)$. At each semantic position, let $r_{\mathrm{dec}}(p)$ be the rank of extended SID prefix $p$ under the decoder's cumulative sequence probability. Using $S_p$ for the items whose SIDs begin with $p$, we give $p$ the rank of its highest-ranked item in $Q(h)$.
\begin{equation}
  r_{\mathrm{item}}(p)=
  \min_{i\in Q(h)\cap S_p}
  r_Q(i),
  \label{eq:prefix-support-rank}
\end{equation}
where $r_Q(i)$ is the rank of $i$ in $Q(h)$. If $Q(h)\cap S_p$ is empty, we set $r_{\mathrm{item}}(p)=\infty$. We then combine the decoder rank and item rank using reciprocal rank fusion~\cite{cormack2009rrf}:
\begin{equation}
  F(p)=\frac{1}{\kappa+r_{\mathrm{dec}}(p)}
  +
  \frac{\mathbf 1[r_{\mathrm{item}}(p)<\infty]}
  {\kappa+r_{\mathrm{item}}(p)}.
  \label{eq:evidence-fusion}
\end{equation}
Here, $\kappa$ is the smoothing constant. Reciprocal-rank fusion places the two rankings on a common scale without learning a calibration between their raw scores~\cite{cormack2009rrf}. We retain the $B$ SID prefixes with the largest $F(p)$ and apply the same rule at every semantic position. The first term favors a prefix ranked highly by the decoder. The second favors a prefix containing a highly ranked item from $Q(h)$. Appendix~\ref{app:isd-details} reports $J$, $B$, $\kappa$, and the history windows.

\subsubsection{\textbf{After SID Completion}}
\label{sec:after-completion} 
Preserving an SID prefix keeps its items available but does not place them near the top of the final recommendation list. Let $r_{\mathrm{SID}}(i)$ be generated item $i$'s rank under the decoder and set $r_Q(i)=\infty$ when $i\notin Q(h)$. We rank the generated items by
\begin{equation}
  F_{\mathrm{item}}(i)
  =\frac{1}{\kappa+r_{\mathrm{SID}}(i)}
  +\frac{\mathbf 1[r_Q(i)<\infty]}{\kappa+r_Q(i)}.
  \label{eq:evidence-fusion-item}
\end{equation}
ISD constructs $Q(h)$ once and uses one decoder pass. It does not change an SID or insert an item whose SID was not generated.

\subsection{Complexity Analysis}

%
We treat the ranked item set $Q(h)$ as a preprocessing input to ISD. Its construction is ranker-dependent and is therefore excluded from the decoding complexity. Let $T_{\mathrm{dec}}$ denote the cumulative cost of the decoder forward passes. Ordinary SID generation considers at most $BV$ prefix extensions at each of $L$ positions, giving runtime
\begin{equation}
  \mathcal{O}\!\left(T_{\mathrm{dec}}+LBV\right).
  \label{eq:standard-decoding-complexity}
\end{equation}

Given $Q(h)$, ISD maps its $J$ items to their $L$ prefixes, checks item support for at most $LBV$ extensions, and orders the $B$ generated items. Its decoding runtime is
\begin{equation}
  \mathcal{O}\!\left(
  T_{\mathrm{dec}}
  + LBV
  + JL
  + B\log B
  \right).
  \label{eq:isd-complexity}
\end{equation}
For a fixed decoding configuration, the item-support budget $J$, SID length $L$, beam width $B$, and maximum number of next-token choices $V$ are constants. ISD therefore has the same asymptotic decoding complexity as ordinary SID generation.

\section{Experiments}
\label{sec:recommendation-evaluation}
\begin{table*}[!t]
  \begin{minipage}[t]{0.485\textwidth}
    \vspace{0pt}
    \centering
    \captionof{table}{Held-out items remaining after each token under alternative SID assignments (\%). }
    \vspace{-5pt}
    \label{tab:alternative-sid-survival}
    \small
    \setlength{\tabcolsep}{0.8pt}
    \renewcommand{\arraystretch}{0.94}
    \begin{tabular*}{\linewidth}{@{\extracolsep{\fill}}lrrrrrr@{}}
      \toprule
      & \multicolumn{3}{c}{Office} & \multicolumn{3}{c}{Scientific} \\
      \cmidrule(lr){2-4}\cmidrule(l){5-7}
      Method & T1 & T2 & T3 & T1 & T2 & T3 \\
      \midrule
      TIGER~\cite{rajput2023recommender}
        & 18.1 & 0.04 & 0.00 & 18.0 & 0.05 & 0.00 \\
      \quad + ISD (statistics)
        & 31.2 & 13.8 & 13.5 & 33.2 & 14.4 & 13.5 \\
      \quad + ISD (SASRec)
        & 33.3 & 14.6 & 13.8 & \textbf{35.9} & \textbf{15.2} & \textbf{13.8} \\
      \quad + ISD (UniSRec)
        & \textbf{34.0} & \textbf{15.3} & \textbf{14.2} & 33.8 & 14.3 & 13.1 \\
      \addlinespace
      LIGER~\cite{yang2024liger}
        & 17.8 & 0.04 & 0.00 & 16.0 & 0.10 & 0.00 \\
      \quad + ISD (statistics)
        & 31.1 & 13.7 & 13.5 & 32.5 & 14.4 & 13.6 \\
      \quad + ISD (SASRec)
        & 33.2 & 14.6 & 13.8 & \textbf{35.4} & \textbf{15.1} & \textbf{13.7} \\
      \quad + ISD (UniSRec)
        & \textbf{33.8} & \textbf{15.3} & \textbf{14.2} & 33.5 & 14.6 & 13.2 \\
      \bottomrule
    \end{tabular*}
  \end{minipage}
  \hfill
  \begin{minipage}[t]{0.485\textwidth}
    \vspace{0pt}
    \centering
    \captionof{table}{Ablation of the two ISD stages. ``Filtering'' acts before beam reduction, while ``ordering'' acts after SID generation.}
    \vspace{-5pt}
    \label{tab:isd-stage-ablation}
    \small
    \setlength{\tabcolsep}{1.6pt}
    \begin{tabular*}{\linewidth}{@{\extracolsep{\fill}}llrrrr@{}}
      \toprule
      & & \multicolumn{2}{c}{Office} & \multicolumn{2}{c}{Scientific} \\
      \cmidrule(lr){3-4}\cmidrule(l){5-6}
      Backbone & ISD stage
        & N@10 & R@10
        & N@10 & R@10 \\
      \midrule
      \multirow{4}{*}{TIGER} & No ISD
        & 0.0949 & 0.1194
        & 0.0802 & 0.1166 \\
      & Filtering only
        & 0.0950 & 0.1194
        & 0.0802 & 0.1167 \\
      & Ordering only
        & 0.1035 & 0.1294
        & 0.0887 & \textbf{0.1250} \\
      & Filtering + ordering
        & \textbf{0.1074} & \textbf{0.1349}
        & \textbf{0.0891} & 0.1235 \\
      \addlinespace
      \multirow{4}{*}{LIGER} & No ISD
        & 0.0983 & 0.1257
        & 0.0880 & 0.1291 \\
      & Filtering only
        & 0.1015 & 0.1323
        & \textbf{0.0934} & \textbf{0.1374} \\
      & Ordering only
        & 0.1026 & 0.1292
        & 0.0910 & 0.1289 \\
      & Filtering + ordering
        & \textbf{0.1081} & \textbf{0.1357}
        & 0.0927 & 0.1266 \\
      \bottomrule
    \end{tabular*}
  \end{minipage}

\end{table*}

In this section, we evaluate Item-Supported Decoding (ISD) on three domains from the Amazon Reviews dataset~\cite{amazon_review}. We use the following research questions to guide our experiments: \textbf{RQ1:} Does ISD improve recommendation performance? \textbf{RQ2:} How much of the improvement comes from supporting SID prefixes during generation, and how much comes from ordering the generated items afterward? \textbf{RQ3:} When the same item is assigned an alternative SID after an equivalent change to its description, can ISD keep that alternative SID available during generation?

\subsection{Datasets and Experimental Details}

\subsubsection{\textbf{Datasets and Splits}} We use the Baby, Office, and Scientific domains from the Amazon Reviews dataset~\cite{amazon_review}. We order each user's interactions chronologically, reserve the final interaction for testing and the preceding interaction for validation, and use the remainder for training. This follows the leave-one-out protocol used by the evaluated sequential and generative recommenders~\cite{kang2018sasrec,rajput2023recommender}. Appendix Table~\ref{tab:recommendation-datasets} reports the processed dataset statistics.

\subsubsection{\textbf{Compared Methods}} We include three sequential item recommenders: SASRec~\cite{kang2018sasrec}, BERT4Rec~\cite{sun2019bert4rec}, and UniSRec~\cite{hou2022unisrec}. SETRec~\cite{lin2025order} avoids left-to-right SID filtering through order-agnostic generation, while ETEGRec~\cite{liu2024etegrec} jointly trains its item tokenizer and generative recommender. TIGER~\cite{rajput2023recommender} and LIGER~\cite{yang2024liger} are the autoregressive SID systems to which we apply ISD. For the item ranking $Q(h)$, we compare the training-statistics score in Eq.~\eqref{eq:behavior-evidence}, which combines first-order transitions and item-based co-consumption~\cite{rendle2010fpmc,sarwar2001item}, with SASRec~\cite{kang2018sasrec} and UniSRec~\cite{hou2022unisrec}. The corresponding rows are denoted ISD (statistics), ISD (SASRec), and ISD (UniSRec).

\subsubsection{\textbf{Evaluation Protocol}} For each held-out interaction, we rank the target item against all other eligible items in the domain and report Recall and NDCG at cutoffs 10 and 20. All methods use the same interaction histories and eligible items. Appendix~\ref{app:recommendation-evaluation} describes the baseline adaptations, model selection, and three-seed evaluation. Appendix~\ref{app:isd-details} reports the fixed ISD settings.

\subsection{Recommendation Results (\textbf{RQ1})}

Table~\ref{tab:recommendation-results} reports the matched comparison and varies the source of the item ranking. Across the three-domain grid, all three sources numerically improve NDCG@10 for both evaluated SID systems. The effect is largest for TIGER on Baby, where training statistics, SASRec, and UniSRec improve NDCG@10 by 18.2\%, 23.9\%, and 31.2\%, respectively. LIGER's Baby gains are smaller, ranging from 1.1\% to 4.5\%. The improvement therefore does not depend on the hand-designed training statistics: rankings supplied by SASRec and UniSRec also improve SID generation without retraining its decoder.

The provider matters for the remaining metrics. With TIGER, training statistics improve NDCG@10 but reduce Recall@20 on Scientific and Baby, whereas SASRec and UniSRec improve all four metrics on all three domains. With LIGER on Baby, the statistics and SASRec gains are modest, while UniSRec gives the largest improvement. These differences are consistent with ISD's role: it preserves and orders the items favored by its provider rather than defining a new source of relevance.

\subsection{Ablation Study (\textbf{RQ2})}

ISD uses the item ranking twice: first to support SID prefixes before beam
filtering and then to order the items whose complete SIDs were generated. We
separate these uses for both SID systems on their matched validation
populations in Table~\ref{tab:isd-stage-ablation}.

\subsection{Generation Under SID Reassignment (\textbf{RQ3})}
Table~\ref{tab:alternative-sid-survival} shows that TIGER and LIGER preserve almost none of the reassigned SID paths through the third semantic token, whereas ISD preserves 13.1--14.2\% across item-ranking methods and domains. Item support therefore helps an existing decoder follow a valid semantic path that differs from the target's original label. Complete retrieval remains zero because the frozen decoder has not learned the reassigned collision-resolution suffix, so ISD improves robustness to semantic reassignment but does not solve exact identity remapping.

The useful stage depends on the downstream head. TIGER receives most of its
gain from ordering, and combining filtering with ordering is strongest on
Office. LIGER already has a dense item head: filtering alone is strongest on
Scientific, while combining both stages gives the highest Office NDCG@10.
Thus filtering and ordering are distinct interventions rather than
interchangeable implementations of the same score.

\subsection{Interpretation and Scope}

The completed comparison supports a limited but useful conclusion. Item evidence can improve an already-trained autoregressive SID system at inference time, and the effect is not specific to one hand-designed ranking rule. ISD does not replace the SID decoder. It changes which SID prefixes remain under the fixed beam and how the generated items are ordered. It also does not address unstable exact item identity from Section~\ref{sec:measurement}.

The comparison averages three runs per trainable method after validation-only
checkpoint selection. We therefore use it to test the design
consequence of Sections~\ref{sec:measurement}--\ref{sec:discretization}, not
to claim that ISD universally improves every SID system or replaces
conventional item recommendation.

\FloatBarrier

\section{Related Work}
\label{sec:related-work}

\textbf{SID-based generative recommendation.}
A generative recommender predicts an item by generating its identifier rather than scoring every item independently. TIGER generates short residual-quantized identifiers from user histories~\cite{rajput2023recommender}. Later methods incorporate collaborative signals, align tokenization with recommendation, or scale SID-based models~\cite{wang2024letter,liu2024etegrec,fu2026diger,kong2025minionerec,sidreasoner2026}. These works improve SID learning or generation. We instead examine how SID construction changes item representations and how its tokens determine which items remain available.

\textbf{SID decoding and item selection.}
Decoding-focused methods address premature removal or sequential token dependence. APAO retrains the decoder with prefix-level objectives~\cite{yu2026apao}. RPG and SETRec generate order-agnostic tokens in parallel~\cite{hou2025rpg,lin2025order}, while LIGER combines SID generation with dense retrieval and ranking~\cite{yang2024liger}. ISD instead keeps the SID assignment and decoder fixed, using an independent item ranking to support prefixes before beam reduction and to order the generated items.

\textbf{Quantization and representation evaluation.}
SID construction builds on product and learned vector quantization~\cite{jegou2011pq,ge2013opq,oord2017vqvae,lee2022rqvae}. Conventional retrieval uses a quantized code to locate candidates and a separate key to identify the item. SID recommendation places both roles in one generated sequence. Our controlled tests therefore distinguish continuous item recognition, neighborhood preservation, and exact SID consistency.

\section{Conclusion}
\label{sec:conclusion}
SIDs provide useful coarse organization, but their exact codes are not determined by semantics alone, and their fine tokens preserve limited local item structure while deciding which items survive generation. ISD reduces this conflict by adding item-level evidence before beam reduction and after generation, improving NDCG@10 by up to 31.2\% across TIGER and LIGER without retraining. SIDs can organize and identify items, but their fine boundaries should not alone control item selection.

\bibliographystyle{ACM-Reference-Format}
\bibliography{paper}

@inproceedings{sun2019bert4rec,
  title={BERT4Rec: Sequential recommendation with bidirectional encoder representations from transformer},
  author={Sun, Fei and Liu, Jun and Wu, Jian and Pei, Changhua and Lin, Xiao and Ou, Wenwu and Jiang, Peng},
  booktitle={Proceedings of the 28th ACM international conference on information and knowledge management},
  pages={1441--1450},
  year={2019}
}

@inproceedings{rajput2023recommender,
title={Recommender Systems with Generative Retrieval},
author={Shashank Rajput and Nikhil Mehta and Anima Singh and Raghunandan Hulikal Keshavan and Trung Vu and Lukasz Heldt and Lichan Hong and Yi Tay and Vinh Q. Tran and Jonah Samost and Maciej Kula and Ed H. Chi and Maheswaran Sathiamoorthy},
booktitle={Thirty-seventh Conference on Neural Information Processing Systems},
year={2023},
url={https://openreview.net/forum?id=BJ0fQUU32w}
}

@inproceedings{amazon_review,
    title = "Justifying Recommendations using Distantly-Labeled Reviews and Fine-Grained Aspects",
    author = "Ni, Jianmo  and
      Li, Jiacheng  and
      McAuley, Julian",
    editor = "Inui, Kentaro  and
      Jiang, Jing  and
      Ng, Vincent  and
      Wan, Xiaojun",
    booktitle = "Proceedings of the 2019 Conference on Empirical Methods in Natural Language Processing and the 9th International Joint Conference on Natural Language Processing (EMNLP-IJCNLP)",
    month = nov,
    year = "2019",
    address = "Hong Kong, China",
    publisher = "Association for Computational Linguistics",
    url = "https://aclanthology.org/D19-1018/",
    doi = "10.18653/v1/D19-1018",
    pages = "188--197"
}

@inproceedings{wang2024letter,
author = {Wang, Wenjie and Bao, Honghui and Lin, Xinyu and Zhang, Jizhi and Li, Yongqi and Feng, Fuli and Ng, See-Kiong and Chua, Tat-Seng},
title = {Learnable Item Tokenization for Generative Recommendation},
year = {2024},
isbn = {9798400704369},
publisher = {Association for Computing Machinery},
address = {New York, NY, USA},
url = {https://doi.org/10.1145/3627673.3679569},
doi = {10.1145/3627673.3679569},
booktitle = {Proceedings of the 33rd ACM International Conference on Information and Knowledge Management},
pages = {2400–2409},
numpages = {10},
location = {Boise, ID, USA},
series = {CIKM '24}
}

@inproceedings{liu2024etegrec,
author = {Liu, Enze and Zheng, Bowen and Ling, Cheng and Hu, Lantao and Li, Han and Zhao, Wayne Xin},
title = {Generative Recommender with End-to-End Learnable Item Tokenization},
year = {2025},
isbn = {9798400715921},
publisher = {Association for Computing Machinery},
address = {New York, NY, USA},
url = {https://doi.org/10.1145/3726302.3729989},
doi = {10.1145/3726302.3729989},
booktitle = {Proceedings of the 48th International ACM SIGIR Conference on Research and Development in Information Retrieval},
pages = {729–739},
numpages = {11},
location = {Padua, Italy},
series = {SIGIR '25}
}

@inproceedings{hou2025rpg,
author = {Hou, Yupeng and Li, Jiacheng and Shin, Ashley and Jeon, Jinsung and Santhanam, Abhishek and Shao, Wei and Hassani, Kaveh and Yao, Ning and McAuley, Julian},
title = {Generating Long Semantic IDs in Parallel for Recommendation},
year = {2025},
isbn = {9798400714542},
publisher = {Association for Computing Machinery},
address = {New York, NY, USA},
url = {https://doi.org/10.1145/3711896.3736979},
doi = {10.1145/3711896.3736979},
booktitle = {Proceedings of the 31st ACM SIGKDD Conference on Knowledge Discovery and Data Mining V.2},
pages = {956–966},
numpages = {11},
location = {Toronto ON, Canada},
series = {KDD '25}
}

@inproceedings{lin2025order,
author = {Lin, Xinyu and Shi, Haihan and Wang, Wenjie and Feng, Fuli and Wang, Qifan and Ng, See-Kiong and Chua, Tat-Seng},
title = {Order-agnostic Identifier for Large Language Model-based Generative Recommendation},
year = {2025},
isbn = {9798400715921},
publisher = {Association for Computing Machinery},
address = {New York, NY, USA},
url = {https://doi.org/10.1145/3726302.3730053},
doi = {10.1145/3726302.3730053},
booktitle = {Proceedings of the 48th International ACM SIGIR Conference on Research and Development in Information Retrieval},
pages = {1923–1933},
numpages = {11},
location = {Padua, Italy},
series = {SIGIR '25}
}

@inproceedings{fu2026diger,
author = {Fu, Junchen and Ge, Xuri and Karatzoglou, Alexandros and Arapakis, Ioannis and Verberne, Suzan and Jose, Joemon M. and Ren, Zhaochun},
title = {Differentiable Semantic ID for Generative Recommendation},
year = {2026},
isbn = {9798400725999},
publisher = {Association for Computing Machinery},
address = {New York, NY, USA},
url = {https://doi.org/10.1145/3805712.3809641},
doi = {10.1145/3805712.3809641},
booktitle = {Proceedings of the 49th International ACM SIGIR Conference on Research and Development in Information Retrieval},
pages = {369–379},
numpages = {11},
location = {Australia},
series = {SIGIR '26}
}

@misc{kong2025minionerec,
      title={MiniOneRec: An Open-Source Framework for Scaling Generative Recommendation}, 
      author={Xiaoyu Kong and Leheng Sheng and Junfei Tan and Yuxin Chen and Jiancan Wu and An Zhang and Xiang Wang and Xiangnan He},
      year={2025},
      eprint={2510.24431},
      archivePrefix={arXiv},
      primaryClass={cs.IR},
      url={https://arxiv.org/abs/2510.24431}, 
}

@misc{decoupledrq2026,
      title={Decoupled Residual Quantization for Robust Semantic IDs in Recommendation}, 
      author={Xuesi Wang and Junjie Wang and Ziliang Wang and Weijie Bian and Guanxing Zhang},
      year={2026},
      eprint={2606.01844},
      archivePrefix={arXiv},
      primaryClass={cs.IR},
      url={https://arxiv.org/abs/2606.01844}, 
}

@inproceedings{penha2025semanticids,
author = {Penha, Gustavo and D'Amico, Edoardo and De Nadai, Marco and Palumbo, Enrico and Tamborrino, Alexandre and Vardasbi, Ali and Lefarov, Max and Lin, Shawn and Heath, Timothy and Fabbri, Francesco and Bouchard, Hugues},
title = {Semantic IDs for Joint Generative Search and Recommendation},
year = {2025},
isbn = {9798400713644},
publisher = {Association for Computing Machinery},
address = {New York, NY, USA},
url = {https://doi.org/10.1145/3705328.3759300},
doi = {10.1145/3705328.3759300},
booktitle = {Proceedings of the Nineteenth ACM Conference on Recommender Systems},
pages = {1296–1301},
numpages = {6},
location = {
},
series = {RecSys '25}
}

@misc{sidreasoner2026,
      title={Reasoning over Semantic IDs Enhances Generative Recommendation}, 
      author={Yingzhi He and Yan Sun and Junfei Tan and Yuxin Chen and Xiaoyu Kong and Chunxu Shen and Xiang Wang and An Zhang and Tat-Seng Chua},
      year={2026},
      eprint={2603.23183},
      archivePrefix={arXiv},
      primaryClass={cs.IR},
      url={https://arxiv.org/abs/2603.23183}, 
}

@article{yang2024liger,
title={Unifying Generative and Dense Retrieval for Sequential Recommendation},
author={Liu Yang and Fabian Paischer and Kaveh Hassani and Jiacheng Li and Shuai Shao and Zhang Gabriel Li and Yun He and Xue Feng and Nima Noorshams and Sem Park and Bo Long and Robert D Nowak and Xiaoli Gao and Hamid Eghbalzadeh},
journal={Transactions on Machine Learning Research},
issn={2835-8856},
year={2025},
url={https://openreview.net/forum?id=jxdnFIsjCb},
note={}
}

@misc{yu2026apao,
title={APAO: Adaptive Prefix-Aware Optimization for Generative Recommendation},
author={Yuanqing Yu and Yifan Wang and Weizhi Ma and Zhiqiang Guo and Min Zhang},
year={2026},
eprint={2603.02730},
archivePrefix={arXiv},
primaryClass={cs.IR},
url={https://arxiv.org/abs/2603.02730}
}

@ARTICLE{jegou2011pq,
  author={Jégou, Herve and Douze, Matthijs and Schmid, Cordelia},
  journal={IEEE Transactions on Pattern Analysis and Machine Intelligence}, 
  title={Product Quantization for Nearest Neighbor Search}, 
  year={2011},
  volume={33},
  number={1},
  pages={117-128},
  doi={10.1109/TPAMI.2010.57}}

@inproceedings{ge2013opq,
author = {Ge, Tiezheng and He, Kaiming and Ke, Qifa and Sun, Jian},
title = {Optimized Product Quantization for Approximate Nearest Neighbor Search},
year = {2013},
isbn = {9780769549897},
publisher = {IEEE Computer Society},
address = {USA},
url = {https://doi.org/10.1109/CVPR.2013.379},
doi = {10.1109/CVPR.2013.379},
booktitle = {Proceedings of the 2013 IEEE Conference on Computer Vision and Pattern Recognition},
pages = {2946–2953},
numpages = {8},
series = {CVPR '13}
}

@inproceedings{oord2017vqvae,
author = {van den Oord, Aaron and Vinyals, Oriol and Kavukcuoglu, Koray},
title = {Neural discrete representation learning},
year = {2017},
isbn = {9781510860964},
publisher = {Curran Associates Inc.},
address = {Red Hook, NY, USA},
booktitle = {Proceedings of the 31st International Conference on Neural Information Processing Systems},
pages = {6309–6318},
numpages = {10},
location = {Long Beach, California, USA},
series = {NIPS'17}
}

@INPROCEEDINGS{lee2022rqvae,
  author={Lee, Doyup and Kim, Chiheon and Kim, Saehoon and Cho, Minsu and Han, Wook-Shin},
  booktitle={2022 IEEE/CVF Conference on Computer Vision and Pattern Recognition (CVPR)}, 
  title={Autoregressive Image Generation using Residual Quantization}, 
  year={2022},
  volume={},
  number={},
  pages={11513-11522},
  doi={10.1109/CVPR52688.2022.01123}}

@inproceedings{ribeiro2020checklist,
    title = "Beyond Accuracy: Behavioral Testing of {NLP} Models with {C}heck{L}ist",
    author = "Ribeiro, Marco Tulio  and
      Wu, Tongshuang  and
      Guestrin, Carlos  and
      Singh, Sameer",
    editor = "Jurafsky, Dan  and
      Chai, Joyce  and
      Schluter, Natalie  and
      Tetreault, Joel",
    booktitle = "Proceedings of the 58th Annual Meeting of the Association for Computational Linguistics",
    month = jul,
    year = "2020",
    address = "Online",
    publisher = "Association for Computational Linguistics",
    url = "https://aclanthology.org/2020.acl-main.442/",
    doi = "10.18653/v1/2020.acl-main.442",
    pages = "4902--4912"
}

@inproceedings{ni2022sentencet5,
    title = "Sentence-T5: Scalable Sentence Encoders from Pre-trained Text-to-Text Models",
    author = "Ni, Jianmo  and
      Hernandez Abrego, Gustavo  and
      Constant, Noah  and
      Ma, Ji  and
      Hall, Keith  and
      Cer, Daniel  and
      Yang, Yinfei",
    editor = "Muresan, Smaranda  and
      Nakov, Preslav  and
      Villavicencio, Aline",
    booktitle = "Findings of the Association for Computational Linguistics: ACL 2022",
    month = may,
    year = "2022",
    address = "Dublin, Ireland",
    publisher = "Association for Computational Linguistics",
    url = "https://aclanthology.org/2022.findings-acl.146/",
    doi = "10.18653/v1/2022.findings-acl.146",
    pages = "1864--1874"
}

@inproceedings{xiao2023cpack,
author = {Xiao, Shitao and Liu, Zheng and Zhang, Peitian and Muennighoff, Niklas and Lian, Defu and Nie, Jian-Yun},
title = {C-Pack: Packed Resources For General Chinese Embeddings},
year = {2024},
isbn = {9798400704314},
publisher = {Association for Computing Machinery},
address = {New York, NY, USA},
url = {https://doi.org/10.1145/3626772.3657878},
doi = {10.1145/3626772.3657878},
booktitle = {Proceedings of the 47th International ACM SIGIR Conference on Research and Development in Information Retrieval},
pages = {641–649},
numpages = {9},
location = {Washington DC, USA},
series = {SIGIR '24}
}

@misc{zhang2025qwen3embedding,
      title={Qwen3 Embedding: Advancing Text Embedding and Reranking Through Foundation Models}, 
      author={Yanzhao Zhang and Mingxin Li and Dingkun Long and Xin Zhang and Huan Lin and Baosong Yang and Pengjun Xie and An Yang and Dayiheng Liu and Junyang Lin and Fei Huang and Jingren Zhou},
      year={2025},
      eprint={2506.05176},
      archivePrefix={arXiv},
      primaryClass={cs.CL},
      url={https://arxiv.org/abs/2506.05176}, 
}

@inproceedings{cormack2009rrf,
author = {Cormack, Gordon V. and Clarke, Charles L A and Buettcher, Stefan},
title = {Reciprocal rank fusion outperforms condorcet and individual rank learning methods},
year = {2009},
isbn = {9781605584836},
publisher = {Association for Computing Machinery},
address = {New York, NY, USA},
url = {https://doi.org/10.1145/1571941.1572114},
doi = {10.1145/1571941.1572114},
booktitle = {Proceedings of the 32nd International ACM SIGIR Conference on Research and Development in Information Retrieval},
pages = {758–759},
numpages = {2},
location = {Boston, MA, USA},
series = {SIGIR '09}
}

@inproceedings{sarwar2001item,
author = {Sarwar, Badrul and Karypis, George and Konstan, Joseph and Riedl, John},
title = {Item-based collaborative filtering recommendation algorithms},
year = {2001},
isbn = {1581133480},
publisher = {Association for Computing Machinery},
address = {New York, NY, USA},
url = {https://doi.org/10.1145/371920.372071},
doi = {10.1145/371920.372071},
booktitle = {Proceedings of the 10th International Conference on World Wide Web},
pages = {285–295},
numpages = {11},
location = {Hong Kong, Hong Kong},
series = {WWW '01}
}

@inproceedings{rendle2010fpmc,
author = {Rendle, Steffen and Freudenthaler, Christoph and Schmidt-Thieme, Lars},
title = {Factorizing personalized Markov chains for next-basket recommendation},
year = {2010},
isbn = {9781605587998},
publisher = {Association for Computing Machinery},
address = {New York, NY, USA},
url = {https://doi.org/10.1145/1772690.1772773},
doi = {10.1145/1772690.1772773},
booktitle = {Proceedings of the 19th International Conference on World Wide Web},
pages = {811–820},
numpages = {10},
location = {Raleigh, North Carolina, USA},
series = {WWW '10}
}

@misc{kang2018sasrec,
      title={Self-Attentive Sequential Recommendation}, 
      author={Wang-Cheng Kang and Julian McAuley},
      year={2018},
      eprint={1808.09781},
      archivePrefix={arXiv},
      primaryClass={cs.IR},
      url={https://arxiv.org/abs/1808.09781}, 
}

@inproceedings{hou2022unisrec,
author = {Hou, Yupeng and Mu, Shanlei and Zhao, Wayne Xin and Li, Yaliang and Ding, Bolin and Wen, Ji-Rong},
title = {Towards Universal Sequence Representation Learning for Recommender Systems},
year = {2022},
isbn = {9781450393850},
publisher = {Association for Computing Machinery},
address = {New York, NY, USA},
url = {https://doi.org/10.1145/3534678.3539381},
doi = {10.1145/3534678.3539381},
booktitle = {Proceedings of the 28th ACM SIGKDD Conference on Knowledge Discovery and Data Mining},
pages = {585–593},
numpages = {9},
location = {Washington DC, USA},
series = {KDD '22}
}

\clearpage
\appendix
\balance

\twocolumn[{%
\begin{@twocolumnfalse}
  \centering
  \includegraphics[width=0.90\textwidth]{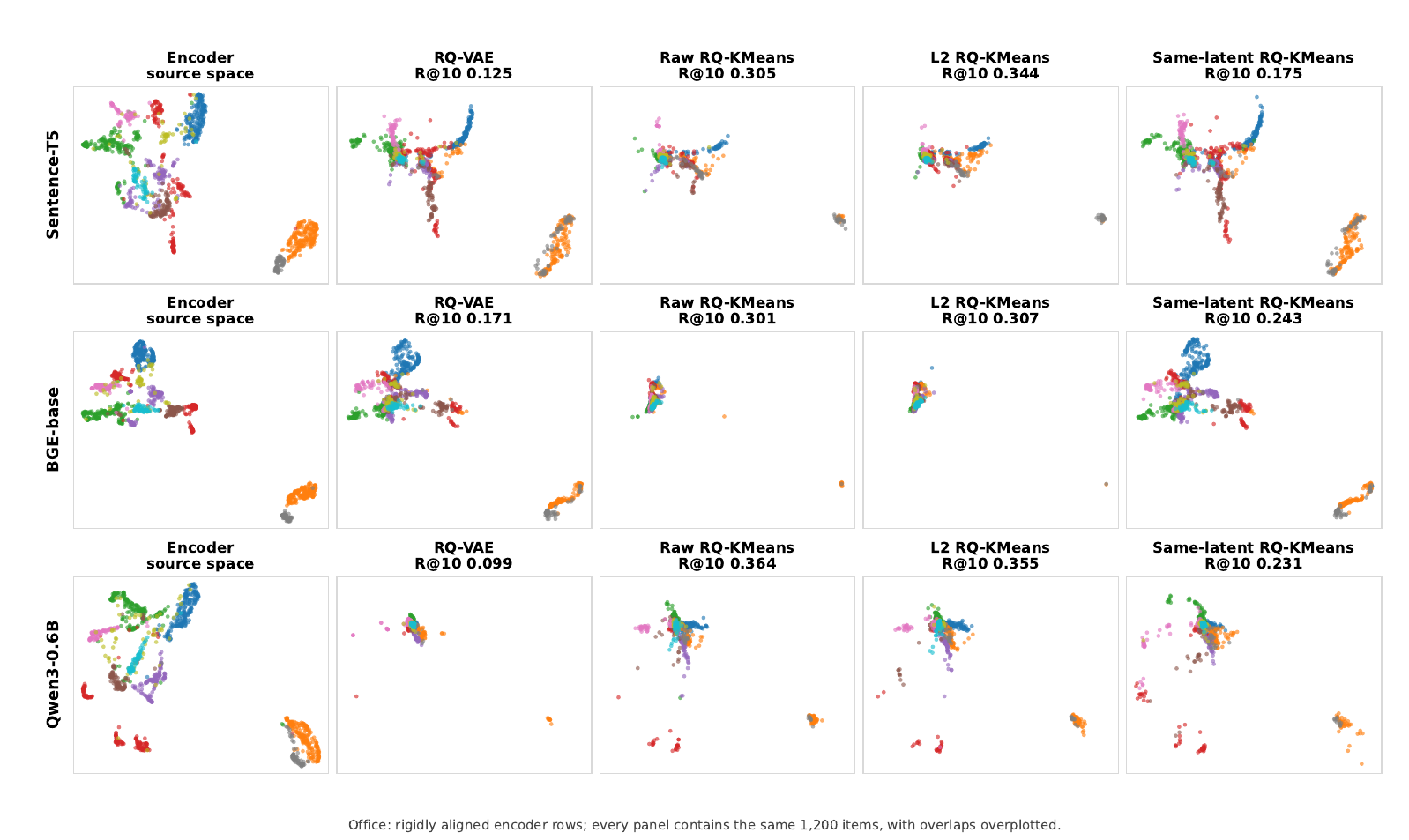}
  \Description{Office panels compare Sentence-T5, BGE-base, and
  Qwen3-Embedding source geometry with RQ-VAE and residual K-Means SID
  reconstructions. Colors denote external product categories, and panel titles
  report encoder-neighborhood Recall at ten.}
  \captionof{figure}{Encoder space and SID-reconstructed geometry on Office. Broad
  category organization remains visible, while fine item neighborhoods change.}
  \label{fig:sid-geometry-crossenc-office}
\end{@twocolumnfalse}
}]

\section{Experimental and Implementation Details}
\label{app:theory}
\small

\subsection{Section 3: Representation Measurements}

\Cref{sec:measurement} compares matched descriptions across Sentence-T5,
BGE, and Qwen encoders. The controlled constructor comparison holds the
Sentence-T5 representation fixed so that only SID construction changes.
\Cref{tab:primary-multiseed} lists the evaluated constructors. Complete-system
checks additionally cover TIGER, RPG, DIGER, LIGER, SIDReasoner, and
MiniOneRec.

\subsubsection{\textbf{Matched-Description Protocol}}

For every item, the source and reordered descriptions contain the same words
and attribute values. Character-level and token-multiset checks verify that the
transformation adds, removes, or changes no item information. An item enters a
constructor comparison only when the implementation exactly reproduces its
stored source SID. This prevents reproduction errors from being counted as SID
changes.

The primary comparison covers Baby, Office, and Scientific. It uses one pinned
Sentence-T5 export per domain and ten frozen constructor seeds. Reported means
average over seeds. Confidence intervals resample both items and seeds. All
nearest-neighbor searches exclude the query item. Encoder neighborhoods use
cosine similarity, while SID neighborhoods use Euclidean distance between
vectors reconstructed from the codebooks.

\subsubsection{\textbf{Additional Local-Structure Checks}}
\label{app:neighborhood}

Exact neighbor overlap is intentionally strict. We therefore also measure
rank-based trustworthiness and continuity, encoder-cosine regret, and category
agreement. Across the evaluated constructors, trustworthiness and continuity
range from 0.950 to 0.994 and cosine regret ranges from 0.043 to 0.104. Thus,
items replacing a missing exact neighbor are often still nearby in the encoder
space. These checks support the main-text conclusion that SIDs retain broad
organization while changing many fine relationships. They do not imply that
the reconstructed neighborhoods are arbitrary.

\subsubsection{\textbf{Geometry Visualizations}}
\label{app:sid-geometry-visualizations}

\Cref{fig:sid-geometry-crossenc-office,fig:sid-geometry-crossenc-scientific}
shows the encoder space and the corresponding SID reconstructions for
Sentence-T5, BGE, and Qwen. Within each row, the projection, items, and category
colors are fixed across constructors. The panels should therefore be compared
within a row. They provide a qualitative view of the quantitative neighborhood
results: broad category organization remains visible, while many local item
relationships change.

\subsubsection{\textbf{Organization Across SID Positions}}
\label{app:position-organization}

Before token \(z_\ell\) is assigned, let
\(p=(z_1,\ldots,z_{\ell-1})\) denote the existing SID prefix. Let \(S_p\)
contain all items whose SIDs begin with \(p\), and let
\(S_{p,v}\subseteq S_p\) contain those assigned next token \(v\). We measure
the separation introduced at position \(\ell\) by
\begin{equation}
I_{p,\ell}
=
\sum_v\frac{|S_{p,v}|}{|S_p|}
\log_2\frac{|S_p|}{|S_{p,v}|}.
\label{eq:resolution-bits}
\end{equation}
This is the Shannon entropy of the empirical next-token distribution. It is
also the expected reduction in the log-size of the remaining item set.

To measure whether the split keeps encoder neighbors together, let
\(N_E^{10}(i)\) be item \(i\)'s self-excluded top-10 encoder neighbors.
For every non-singleton child group, we compute the fraction of these neighbors
that receive the same next token as \(i\). We compare the observed fraction
with 200 randomized assignments within the same parent prefix. Each
randomization preserves the parent membership, the number and size of child
groups, and their item-popularity composition. The excess over this randomized
baseline is the source-neighbor organization introduced by the observed split.

We aggregate separation and excess organization across prefixes using the
number of items under each prefix as weights. Their ratio gives the
source-neighbor organization per bit reported in \Cref{fig:token-roles}.
Parent-balanced and common-support variants give the same qualitative pattern.
Uncertainty is estimated by bootstrapping parent prefixes within constructor
seeds. The collision-resolution suffix is excluded throughout.

\clearpage
\twocolumn[{%
\begin{@twocolumnfalse}
  \centering
  \includegraphics[width=0.90\textwidth]{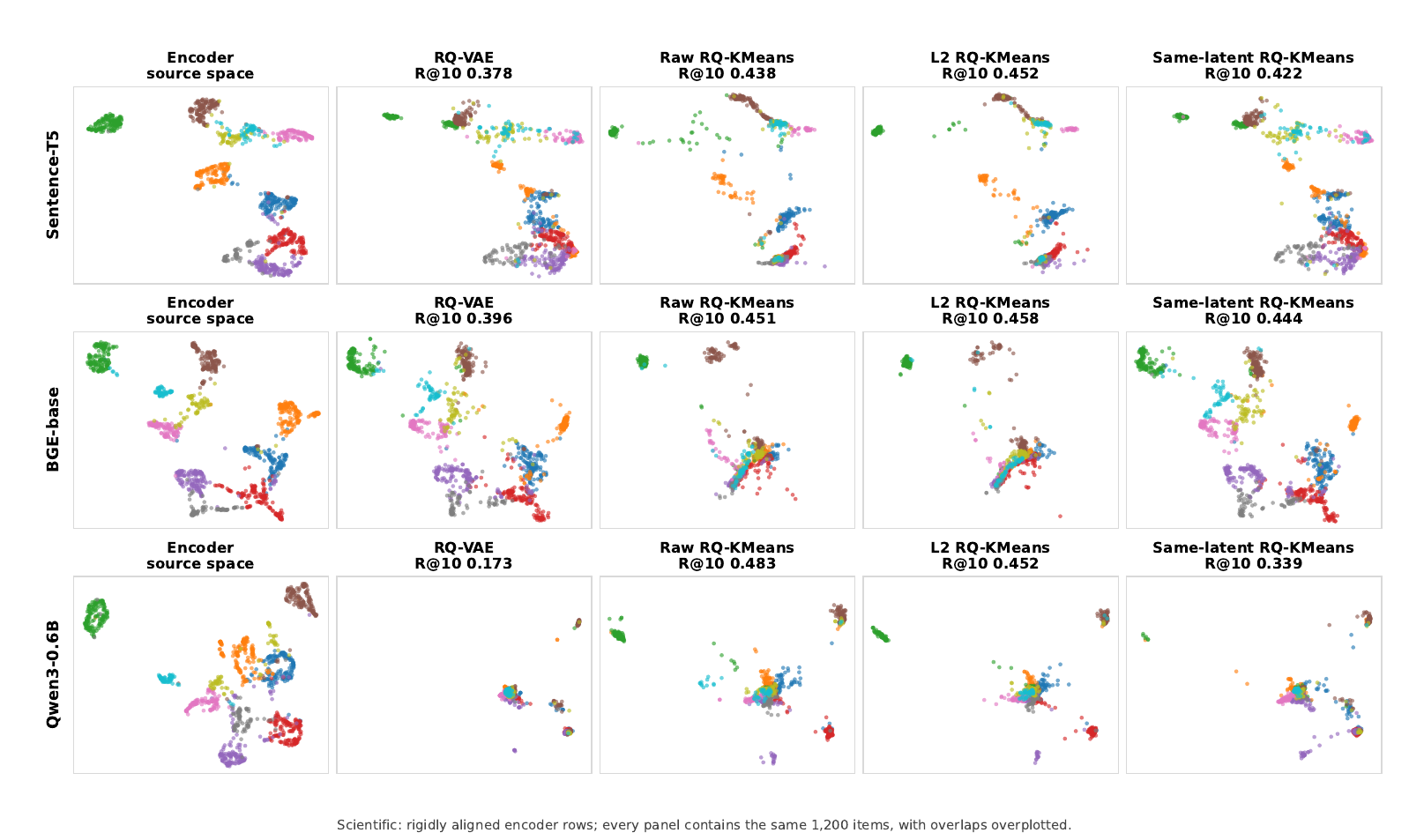}
  \Description{Scientific panels compare Sentence-T5, BGE-base, and
  Qwen3-Embedding source geometry with RQ-VAE and residual K-Means SID
  reconstructions. Colors denote external product categories, and panel titles
  report encoder-neighborhood Recall at ten.}
  \captionof{figure}{Encoder space and SID-reconstructed geometry on Scientific. Broad
  category organization remains visible, while fine item neighborhoods change.}
  \label{fig:sid-geometry-crossenc-scientific}
\end{@twocolumnfalse}
}]

\subsection{Section 4: Target Filtering}
\label{app:semantic-behavior}
\label{app:rankable-target-protocol}

\Cref{sec:discretization} follows each held-out target through SID generation.
The primary diagnostic uses TIGER-style decoders with RQ-VAE SIDs, beam width
50, and decoder seeds 42--44. The figure also reports the corresponding
validation-selected LIGER result. At each semantic position, a target remains
available if at least one retained SID prefix agrees with the target's assigned
prefix through that position.

\subsubsection{\textbf{Top-\(k\) Target Population}}

A separate sequential item scorer is trained only on training interactions.
Before SID generation, it ranks every eligible unconsumed item from the
observed history. We retain histories for which this scorer places the held-out
target among its top \(k\) items. This isolates targets that could plausibly
appear near the top of a recommendation list before SID filtering. The
reported experiment uses \(k=10\). Its fixed population contains 1,198 Office
histories with 345 distinct targets and 536 Scientific histories with 214
distinct targets. All position-wise comparisons use these same histories.

The unweighted metric \(K_\ell\) in \Cref{sec:target_rank} counts the fraction
of this population still available after semantic position \(\ell\). For a
rank-sensitive check, let \(r_h\leq k\) be the target's rank under the
unfiltered item scorer and set \(w_h=1/\log_2(1+r_h)\). We report
\begin{equation}
V_\ell=
\frac{\sum_{h\in\mathcal H_k}w_h
\mathbf 1[i^\star(h)\in\mathcal R_\ell]}
{\sum_{h\in\mathcal H_k}w_h}.
\end{equation}
This gives more weight to the removal of targets that the independent scorer
would place closer to first.

\subsubsection{\textbf{Full-Information Intervention}}

The diagnostic intervention scores every eligible item beneath every possible
next token before beam reduction. For a candidate extension \((p,v)\), its item
score is the largest score among unconsumed items whose SIDs begin with
\((p,v)\). We standardize this value and the decoder score across the possible
next tokens, combine them with a weight selected on validation data, and then
apply the original beam budget. A zero-variance signal is set to zero.
Validation selects weight \(\alpha=0.75\).

The held-out target is never identified, inserted, or restored. However, the
intervention requires scoring every eligible item before every token decision.
It is therefore used only to test whether item-level evidence can prevent
removal. Section~\ref{sec:repair-discussion} evaluates the efficient
approximation.

\subsection{Section 5: Item-Supported Decoding}
\label{app:isd-details}

\subsubsection{\textbf{ISD Settings}}

The ranked item set \(Q(h)\) contains \(J=50\) items and is constructed once
per history. The SID decoder retains \(B=50\) prefixes at each of three semantic
positions and returns at most 50 items. Reciprocal-rank fusion uses
\(\kappa=60\). These settings are fixed across datasets, SID systems, and item
ranking providers.

Training-statistics scores combine last-item transitions with co-consumption
over the ten most recent items. SASRec and UniSRec also provide \(Q(h)\).

\subsubsection{\textbf{Training-Statistics Normalization}}

Let \(E_h\) be the eligible unconsumed items for history \(h\), and let \(u_i\)
be one of the two training-statistics signals for item \(i\). We compute
\begin{equation}
  \mu_h(u)=\frac{1}{|E_h|}\sum_{j\in E_h}u_j,
  \qquad
  \sigma_h(u)=
  \sqrt{\frac{1}{|E_h|}\sum_{j\in E_h}\bigl(u_j-\mu_h(u)\bigr)^2}.
\end{equation}
The standardized value used in Eq.~\eqref{eq:behavior-evidence} is
\begin{equation}
  z_h(u_i)=
  \begin{cases}
    \bigl(u_i-\mu_h(u)\bigr)/\sigma_h(u), & \sigma_h(u)>0,\\
    0, & \sigma_h(u)=0.
  \end{cases}
\end{equation}

\subsection{Section 6: Recommendation Evaluation}
\label{app:recommendation-evaluation}

\begin{table}[H]
  \centering
  \caption{Dataset statistics after preprocessing.}
  \label{tab:recommendation-datasets}
  \small
  \setlength{\tabcolsep}{5pt}
  \begin{tabular*}{\columnwidth}{@{\extracolsep{\fill}}lrrr@{}}
    \toprule
    Dataset & Items & Users & Training interactions \\
    \midrule
    Scientific & 5,334 & 11,024 & 54,958 \\
    Office & 27,965 & 101,478 & 597,310 \\
    Baby & 7,050 & 19,445 & 121,902 \\
    \bottomrule
  \end{tabular*}
\end{table}

\subsubsection{\textbf{Data and Model Selection}}

Interactions are ordered chronologically. The last interaction is held out for
testing, the preceding interaction is used for validation, and the remainder
forms the training set. All compared methods use the same histories, eligible
items, and consumed-item mask. Model checkpoints and ISD providers are selected
using validation NDCG@10.

\subsubsection{\textbf{Evaluation Protocol}}

We retain each baseline's model architecture, training objective, and
recommended hyperparameters while using the common data splits and item masks.
We train each method with three random seeds, select checkpoints on validation
data, and report the arithmetic mean across seeds. The test set is evaluated
only after model selection. For each held-out interaction, we rank the target
against all eligible items and report Recall and NDCG at cutoffs 10 and 20.
Every method returns at most 50 items.


\end{document}